\newif\ifarxiv
\arxivtrue % or
\ifarxiv
	\documentclass[12pt]{article}
	\usepackage[total={6.5in, 9in}]{geometry}
	\usepackage{tgtermes}
	\usepackage[T1]{fontenc}
	\usepackage[urw-garamond]{mathdesign}
\else
	\documentclass[ba]{imsart}
	\pubyear{0000}
	\volume{00}
	\issue{0}
	\doi{0000}
	\firstpage{1}
	\lastpage{1}
\fi

\usepackage{enumitem}   
\usepackage{amsthm}
\usepackage{amsmath}
\usepackage{natbib}
\usepackage[colorlinks,citecolor=blue,urlcolor=blue,filecolor=blue,backref=page]{hyperref}
\usepackage{graphicx, float,wrapfig}
\usepackage{amsfonts}
\usepackage{marginnote}

\newcommand{\enorm}[1]{\Vert #1 \Vert_2}
\newcommand{\defeq}{\mathrel{\mathop:}=}

\graphicspath{{./fig/}}

\begin{document}
\ifarxiv
\author{
	Nicholas G. Polson\\
	\textit{Booth School of Business}\\
	\textit{University of Chicago}\footnote{Polson is Professor of Econometrics and Statistics
		at the Chicago Booth School of Business. email: ngp@chicagobooth.edu. Sokolov is an assistant professor at George Mason University, email: vsokolov@gmu.edu}\\
	Vadim O. Sokolov\\
	\textit{Systems Engineering and Operations Research}\\
	\textit{George Mason University}\\
}
\fi
\ifarxiv
\title{Deep Learning: A Bayesian Perspective}%\thanksref{T1}}
\date{First Draft: May 2017\\
	This Draft: November 2017
}
\maketitle
\fi

\ifarxiv
\begin{abstract}
	\noindent
Deep learning is a form of machine learning for nonlinear high dimensional pattern matching and prediction.  By taking a  Bayesian probabilistic perspective, we provide a number of insights into more efficient algorithms for optimisation and hyper-parameter tuning. Traditional high-dimensional data reduction techniques, such as  principal component analysis (PCA), partial least squares (PLS), reduced rank regression (RRR), projection pursuit regression (PPR) are all shown to be shallow learners. Their deep learning counterparts exploit multiple deep layers of data reduction which provide predictive performance gains. Stochastic gradient descent (SGD) training optimisation and Dropout (DO) regularization provide estimation and variable selection. Bayesian regularization is central to finding  weights and connections in networks to optimize the predictive bias-variance trade-off. To illustrate our methodology, we provide an analysis of international bookings on Airbnb. Finally, we conclude with directions for future research.
\end{abstract}
\fi

\section{Introduction}
Deep learning (DL) is a form of machine learning that uses hierarchical abstract layers of latent variables to perform pattern matching and prediction. Deep learners are  probabilistic predictors where the conditional mean is a stacked generalized linear model (sGLM).  The current interest in DL stems from its remarkable success in a wide range of applications, including  Artificial Intelligence (AI) 
\citep{noauthor_deepmind_nodate, kubota_artificial_2017, esteva2017dermatologist}, image processing \citep{simonyan2014}, learning in games \citep{alphago}, neuroscience \citep{poggio_deep_2016}, energy conservation \citep{noauthor_deepmind_nodate}, and skin cancer diagnostics \citep{kubota_artificial_2017, esteva2017dermatologist}. \cite{schmidhuber2015deep} provides a comprehensive historical survey of deep learning and their applications.   

Deep learning is designed for massive data sets with many high dimensional input variables. For example, Google's translation algorithm \citep{sutskever2014sequence} uses $\sim$ 1-2 billion parameters and very large dictionaries. Computational speed is essential, and automated differentiation and matrix manipulations are available  on {\tt TensorFlow} \cite{tensorflow2015-whitepaper}. Baidu successfully deployed  speech recognition systems \citep{amodei2016deep} with  an extremely large deep learning model with over 100 million parameters, 11 layers and almost 12 thousand hours of speech for training. DL is an algorithmic approach rather than probabilistic in its nature, see \cite{breiman_statistical_2001}  for the merits of both approaches.

Our approach is Bayesian and probabilistic. We view the theoretical roots of DL in Kolmogorov's representation of a multivariate response surface as a superposition of univariate activation functions applied to an affine transformation of the input variable \citep{kolmogorov_representation_1963}. An affine transformation of a vector is a weighted sum of its elements (linear transformation) plus an offset constant (bias). Our Bayesian perspective on DL leads to  new avenues of research including faster stochastic algorithms, hyper-parameter tuning, construction of  good predictors, and model interpretation. 

On the theoretical side, we show how DL exploits a Kolmogorov's ``universal basis''. By construction, deep learning models are very flexible and gradient information can be efficiently calculated for a variety of architectures. On the empirical side, we show that the advances in DL are due to:
\begin{enumerate}[label=(\roman*)]
	\item New activation (a.k.a. link) functions, such as rectified linear unit ($\text{ReLU}(x) = \max(0,x)$), instead of sigmoid function
	\item Depth of the architecture and dropout as a variable selection technique
	\item Computationally efficient routines to train and evaluate the models as well as accelerated computing via graphics processing unit (GPU) and tensor processing unit (TPU)
	\item Deep learning has very well developed computational software where pure MCMC is too slow.
\end{enumerate}
 To illustrate DL, we provide an analysis of a dataset from Airbnb on first time international bookings. Different statistical methodologies can then be compared, see \cite{airbnb} and \cite{ripley1994neural} who provides a comparison of traditional statistical methods with neural network based approaches for classification. 

The rest of the paper is outlined as follows. Section 1.1 provides a review of deep learning. Section 2 provides a Bayesian probabilistic interpretation of many traditional statistical techniques (PCA, PCR, SIR, LDA) which are shown to be ``shallow learners'' with two layers.
Much of the recent success in DL applications has been achieved by including deeper layers and these gains pass over to traditional statistical models. Section 3 provides heuristics on why Bayes procedures provide good predictors in high dimensional data reduction problems.
Section 4 describes how to train, validate and test deep learning models. We provide computational details associated with stochastic gradient descent (SGD).
Section 5 provides an application to bookings data from the Airbnb website.  Finally, Section 6 concludes with directions for future research.

\subsection{Deep Learning}
Machine learning finds a predictor of an output $Y$ given a high dimensional input $X$. A learning machine is an input-output mapping, $Y= F(X)$, where the input space is high-dimensional,
$$
Y = F(X ) \; \; {\rm where} \; \; X = ( X_1 , \ldots , X_p ).
$$
The output $Y$ can be continuous, discrete or mixed.
For a classification problem, we need to learn $ F :X \rightarrow Y$, where $Y \in \{ 1 , \ldots , K \} $ indexes categories. A predictor is denoted by $ \hat{Y}(X)$.

To construct a multivariate function, $F(X)$, we start with building blocks of hidden layers.
Let $f_1,\ldots,f_l$ be univariate activation functions. A  semi-affine activation rule is given by
\[
f_l^{W,b} = f_l\left(\sum_{j=1}^{N_j}W_{ij}z_j + b_l\right) %= f_l(W_lz_l + b_l).
\]
Here $W$ and $z$ are the weight matrix and inputs of the $l$th layer. 

Our deep predictor, given the number of layers $L$, then becomes the composite map
\begin{equation*}
\hat{Y}(X) \defeq \left ( f_1^{W_1,b_1} \circ \ldots \circ f_L^{W_L,b_L} \right ) ( X)\,.\label{DLComp}
\end{equation*}
Put simply, a high dimensional mapping, $F$, is modeled via the superposition of univariate semi-affine functions. Similar to a classic basis decomposition, the deep approach uses univariate activation functions to decompose a high dimensional $X$. To select the number of hidden units (a.k.a neurons), $N_l$, at each layer we will use a stochastic search technique known as dropout.

The offset vector is essential. For example, using $f(x) = \sin(x)$ without bias term $b$ would not allow to recover an even function like $\cos(x)$. An offset element (e.g. $\sin(x+ \pi/2) = \cos(x)$) immediately corrects this problem.

Let $ Z^{(l)} $ denote the $l$-th layer, and so $ X = Z^{(0)}$.
The final output is the response $Y$, which can be numeric or categorical.
A deep prediction rule is then
\begin{align*}
Z^{(1)} & = f^{(1)} \left ( W^{(0)} X + b^{(0)} \right ),\\
Z^{(2)} & = f^{(2)} \left ( W^{(1)} Z^{(1)} + b^{(1)} \right ),\\
  & \ldots\\
Z^{(L)} & = f^{(L)} \left ( W^{(L-1)} Z^{(L-1)} + b^{(L-1)} \right ),\\
\hat{Y} (X) & = W^{(L)} Z^{(L)} + b^{(L)}\,.
\end{align*}
Here, $W^{(l)} $ are weight matrices, and $b^{(l)} $ are threshold or activation levels.
Designing a good predictor depends crucially on the choice of univariate activation functions $ f^{(l)} $.
Kolmogorov's representation requires only two layers in principle. 
\cite{vitushkin_linear_1967} prove the remarkable fact that a discontinuous link is required at the second layer even though the multivariate function is continuous. 
Neural networks (NN) simply approximate a univariate function as mixtures of sigmoids, typically with an exponential number of neurons, which does not generalize well.
They can simply be viewed as projection pursuit regression $F(X) = \sum_{i=1}^N g_i ( W X + b ) )$ with the only difference being that in a neural network the nonlinear link functions, are parameter dependent and learned from training data.

\begin{figure}[H]
	\begin{tabular}{ccc}
		\includegraphics[width=0.18\textwidth]{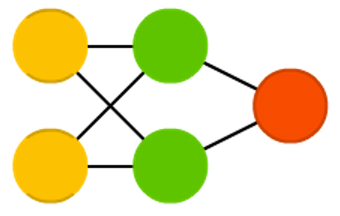} & \includegraphics[width=0.18\textwidth]{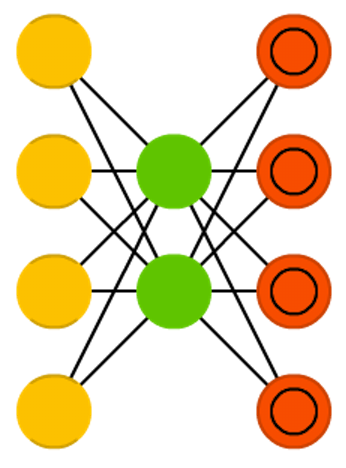} & \includegraphics[width=0.28\textwidth]{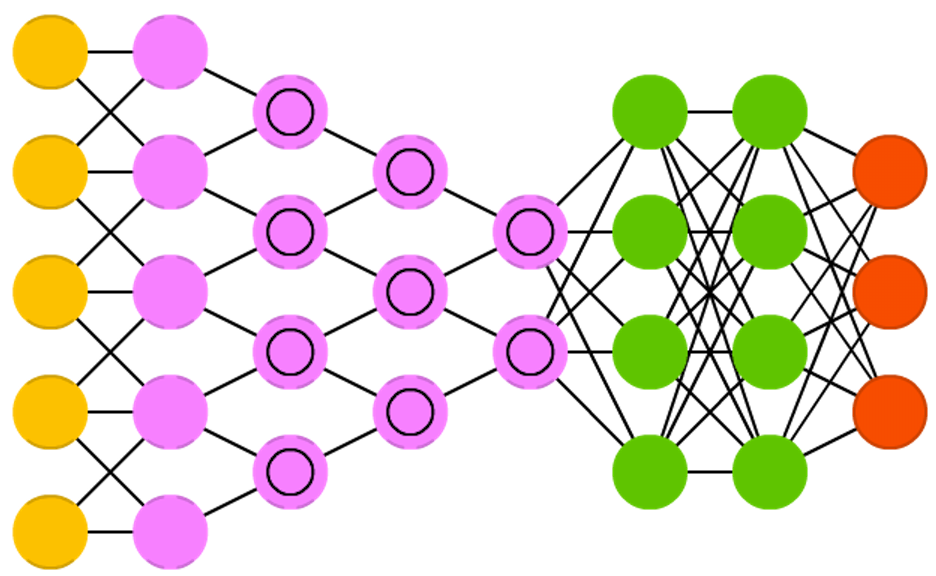}\\ 
		Feed forward & Auto-encoder & Convolution\\
		\includegraphics[width=0.23\textwidth]{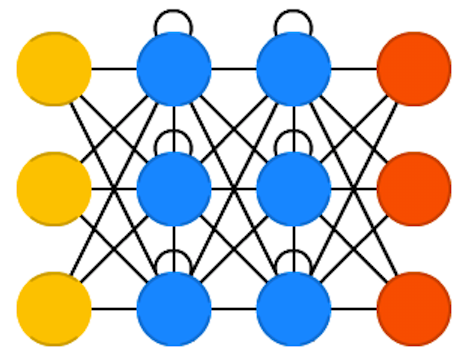} & \includegraphics[width=0.23\textwidth]{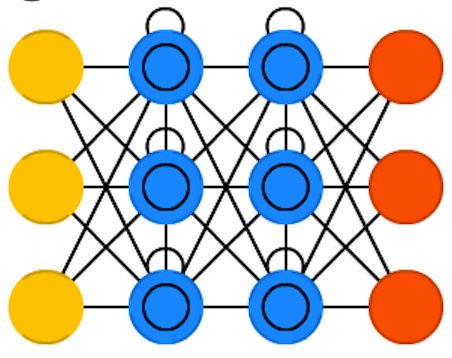}&
		\includegraphics[width=0.23\textwidth]{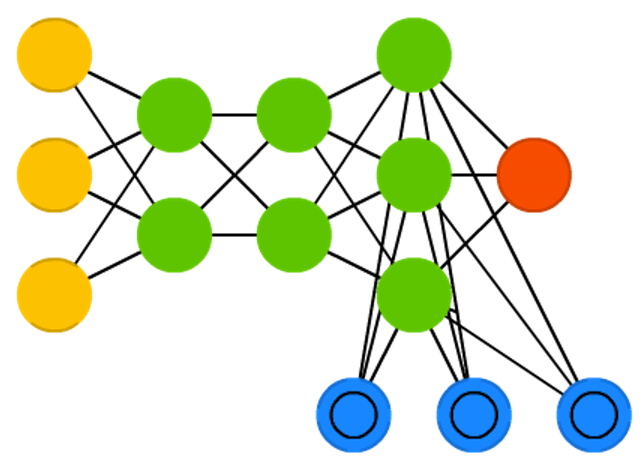}\\
		Recurrent & Long / short term memory & Neural Turing machines
	\end{tabular}
	\caption{Most commonly used deep learning architectures. Each circle is a neuron which calculates a weighted sum of an input vector plus bias and applies a non-linear function to produce an output. Yellow and red colored neurons are input-output cells correspondingly. Pink colored neurons apply wights inputs using a kernel matrix. Green neurons are hidden ones. Blue neurons are recurrent ones and they append its values from previous pass to the input vector. Blue neuron with circle inside a neuron corresponds to a memory cell. Source:	\url{http://www.asimovinstitute.org/neural-network-zoo}.}
	\label{fig:arch}
\end{figure}

Figure \ref{fig:arch} illustrates a number of commonly used structures; for example, feed-forward architectures, auto-encoders, convolutional, and neural Turing machines. Once you have learned the dimensionality of the weight matrices which are non-zero, there's an implied network structure. 

Recently deep architectures (indicating non-zero weights) include convolutional neural networks (CNN), recurrent NN (RNN), long short-term memory (LSTM), and neural Turing machines (NTM).
\cite{pascanu_how_2013} and \cite{montufar_when_2015} provide results on the advantage of representing some functions compactly with deep layers. 
\cite{poggio_deep_2016} extends theoretical results on when deep learning can be exponentially better than shallow learning. 
\cite{bryant_analysis_2008} implements \cite{sprecher_survey_1972} algorithm to estimate the non-smooth inner link function.  In practice, deep layers allow for smooth activation functions to provide ``learned''  hyper-planes which find the underlying complex interactions and regions  without having to see an exponentially large number of training samples. 
 
\section{Deep Probabilistic Learning}
Probabilistically, the output $Y$ can be viewed as a random variable being generated by a probability 
model $p(Y| Y^{W,b}(X))$. Given $\hat{W},\hat{b}$, the negative log-likelihood defines $ \mathcal{L} $ as 
$$\mathcal{L}(Y, \hat{Y} ) = - \log p( Y| Y^{ \hat{W} , \hat{b} } (X) ). $$
The $L_2$-norm, $ \mathcal{L}( Y_i , \hat{Y}( X_i)) = \|Y_i - \hat{Y}( X_i)\|^2_2  $ is traditional least squares, and negative cross-entropy loss is 
$\mathcal{L}( Y_i , \hat{Y}( X_i)) = -\sum_{i=1}^n Y_i \log \hat{Y} ( X_i )$ for multi-class logistic classification. The procedure to obtain estimates $\hat{W},\hat{b}$ of the deep learning model parameters is described in  Section \ref{sec:algorithms}.	

To control the predictive bias-variance trade-off we add a  regularization term and optimize
$$\mathcal{L}_{\lambda}(Y, \hat{Y} ) = - \log p( Y| Y^{ \hat{W} , \hat{b} } (X) )- \log p( \phi(W, b) \mid \lambda).$$
Probabilistically this is a negative log-prior distribution over parameters, namely
\begin{align*}
- \log p( \phi(W, b) \mid \lambda) & =  \lambda \phi(W,b),\\
p( \phi(W, b) \mid \lambda ) & \propto \exp ( - \lambda \phi(W,b)).
\end{align*}
Deep predictors are regularized maximum a posteriori (MAP) estimators, where
\begin{align*}
p( W, b | D ) & \propto  p( Y| Y^{W ,b } (X) ) p( W, b) \\
& \propto  \exp \left ( - \log p( Y| Y^{W ,b } (X) ) - \log p( W, b) \right ).
\end{align*}
Training requires the solution of a highly nonlinear optimization  
$$
\hat{Y} \defeq Y^{ \hat{W} , \hat{b} } (X) \; \; {\rm where} \; \; ( \hat{W} , \hat{b} ) \defeq 
{\rm arg \; max}_{W,b} \; \log p( W, b | D),
$$
and the log-posterior is optimised given the training data, $D = \{ Y^{(i)} , X^{(i)} \}_{i=1}^T$ with
\[ 
- \log p( W, b | D ) = \sum_{i=1}^T \mathcal{L}( Y^{(i)} , Y^{W,b} (X^{(i)} ) ) + \lambda \phi( W, b ).
\]
Deep learning has the key property that $ \nabla_{W,b}  \log p( Y| Y^{W ,b } (X) ) $ is computationally inexpensive to evaluate using tensor methods 
for very complicated architectures and fast implementation on large datasets. {\tt TensorFlow} and {\tt TPUs} provide a state-of-the-art framework for a plethora of architectures.
From a statistical perspective, one caveat is that the posterior is highly multi-modal and providing good hyper-parameter tuning can be expensive.
This is clearly a fruitful area of research for state-of-the-art stochastic Bayesian MCMC algorithms to provide more efficient algorithms.
For shallow architectures,  the alternating direction method of multipliers (ADMM) is an efficient solution to the optimization problem.

\subsection{Dropout for Model and Variable Selection}
Dropout is a model selection technique designed to avoid over-fitting in the training process. This is achieved by removing input dimensions in $X$ randomly with a given probability $p$. 
It is instructive to see how this affects the underlying loss function and optimization problem. For example, suppose that we wish to minimise MSE,  $\mathcal{L}(Y,\hat{Y})=\|Y-\hat{Y}\|^2_2$, then, when marginalizing over the randomness, we have a new objective
$$
{\rm arg \; min}_W \; \mathbb{E}_{ D \sim {\rm Ber} (p) } \Vert Y - W ( D \star X ) \Vert^2_2\,,
$$
Where $ \star$ denotes the element-wise product. It is equivalent to, with $ \Gamma = ( {\rm diag} ( X^\top X) )^{\frac{1}{2}} $
$$
{\rm arg \; min}_W \;  \Vert Y - p W X \Vert^2_2 + p(1-p) \Vert \Gamma W \Vert^2_2.
$$

Dropout then is simply  Bayes ridge regression with a $g$-prior as an objective function. This reduces the likelihood of over-reliance on small sets of input data in training,
see \cite{hinton_reducing_2006} and \cite{srivastava_dropout:_2014}. Dropout can also be viewed as the optimization version of the traditional spike-and-slab prior, which has proven so popular in Bayesian model averaging. For example, in a simple model with one hidden layer, we replace the network
\begin{align*}
Y_i^{(l)} & =  f ( Z_i^{(l)} ), \\
Z_i^{(l)} & = W_i^{(l)} X^{(l)}  + b_i^{(l)},
\end{align*}
with the dropout architecture
\begin{align*}
D_i^{(l)} & \sim \text{Ber} (p), \\
\tilde{Y}_i^{(l)} & = D^{(l)} \star X^{(l)}, \\
Y_i^{(l)} & =  f ( Z_i^{(l)} ), \\
Z_i^{(l)} & = W_i^{(l)} X^{(l)}  + b_i^{(l)}.
\end{align*}
In effect, this replaces the input $X$ by $ D \star X $, where $D$ is a matrix of independent  $\text{Ber}(p)$ distributed random variables. 

Dropout also regularizes the choice of the number of hidden units in a layer. This can be achieved if we drop units of the hidden rather than the input layer and then establish which probability $p$ gives the best results. It is worth recalling though, as we have stated before, one of the dimension reduction properties of a network structure
is that once a variable from a layer is dropped, all terms above it in the network also disappear.

\subsection{Shallow Learners}

Almost all shallow data reduction techniques can be viewed as consisting of a low dimensional auxiliary
variable $Z$ and a prediction rule specified by a
composition of functions 
\[
\hat{Y} = f_1^{W_1,b_1} (f_2( W_2X +b_2)\big) 
= f_1^{W_1,b_1}(Z),\,\text{ where $Z:=f_2(W_2X+b_2)$. }
\]
The problem of high dimensional data reduction is to find the $Z$-variable and to estimate the layer functions $(f_1, f_2)$ correctly. In the layers, we want to uncover the low-dimensional $Z$-structure, in a way that does not disregard information about predicting the output $Y$.

Principal component analysis (PCA), partial least squares (PLS), reduced rank regression (RRR), linear discriminant analysis (LDA), project pursuit regression (PPR), 
and logistic regression are all shallow learners. 
\cite{mallows1973some} provides an interesting perspective on how Bayesian shrinkage provides good predictors in regression settings.
\cite{frank1993statistical} provide excellent discussions of PLS and why Bayesian shrinkage methods provide good predictors.
 \cite{wold1956causal}, \cite{diaconis_nonlinear_1984}, \cite{ripley1994neural}, \cite{cook_fisher_2007,hastie_elements_2016} provide further discussion of dimension reduction techniques.
Other connections exists for Fisher's Linear Discriminant classification rule, which is simply fitting $ H( wX + b)$, where $H$ is a Heaviside function.  \cite{polson_proximal_2015} provide a Bayesian version of support vector machines (SVMs) and a comparison with logistic regression for classification. 

PCA reduces $X$ to $f_2(X)$ using a
singular value decomposition of the form
\begin{equation}
Z =  f_2(X) = W^\top X + b\,,\label{PCA_eq}
\end{equation}
where the columns of the weight matrix $W$ form an orthogonal basis for directions of greatest variance (which is in effect an eigenvector problem). 

Similarly PPR reduces $X$ to $f_2(X)$ by setting
$$
Z=f_2(X) = \sum^{N_1}_{i=1}f_i(W_{i1}X_1 + \ldots + W_{ip}X_p)\,.
$$

\noindent{\bf Example:} 
Interaction terms, $ x_1 x_2 $ and $ (x_1 x_2)^2 $, and max functions, $ \max (x_1, x_2) $  can be expressed as nonlinear functions of semi-affine combinations.
Specifically,  
$$
x_1x_2 = \frac{1}{4} ( x_1+x_2 )^2 - \frac{1}{4} (x_1-x_2)^2
$$
$$
\max(x_1,x_2) = \frac{1}{2} | x_1+x_2 | + \frac{1}{2} | x_1-x_2 |
$$
$$
(x_1x_2)^2 = \frac{1}{4} ( x_1+x_2 )^4 + \frac{7}{4 \cdot 3^3} (x_1-x_2)^4 - \frac{1}{2 \cdot 3^3} ( x_1+ 2 x_2)^4 - \frac{2^3}{3^3} ( x_1 + \frac{1}{2} x_2 )^4
$$
\cite{diaconis1981generating} provide further discussion for Projection Pursuit Regression, where the network uses a layered model of the form
$ \sum_{i=1}^N f ( w_i^\top X ) $. \cite{diaconis1998consistency} provide an ergodic view of composite iterated functions, a precursor to the use of multiple layers of single operators that
can model complex multivariate systems.
\cite{sjoberg1995nonlinear} provide the approximation theory for composite functions.

\vspace{1em}
\noindent{\bf Example: } 
Deep ReLU architectures can be viewed as Max-Sum networks via the following simple identity. 
Define $ x^+ = \max(x,0) $. Let $ f_x ( b ) = ( x + b )^+ $ where $b$ is an offset. Then $ ( x + y^+ )^+ = \max ( 0 , x , x+y ) $. This is generalized
in \cite{feller_introduction_1971} (p.272) who shows by induction that 
$$
( f_{x_1} \circ \ldots \circ f_{x_k} ) (0) = ( x_1 + ( x_2 + \ldots + ( x_{k-1} + x_k^+ )^+ )^+
= \max_{1 \leq j \leq k} ( x_1 + \ldots + x_j )^+ 
$$
A composition or convolution of $ \max $-layers is then a one layer max-sum network.

\subsection{Stacked Auto-Encoders}

Auto-encoding is an important data reduction technique.
An auto-encoder is a deep learning architecture designed to replicate $X$ itself, namely $X=Y$, via a \emph{bottleneck} structure.
This means we select a model $F^{W,b} (X) $ which aims to concentrate the information required to recreate $X$. See \cite{heaton2017deep} for an application to smart indexing in finance.
Suppose that we have $ N $ input vectors $ X = \{ x_1 , \ldots , x_N \} \in \mathbb{R}^{M\times N} $ and
$ N $ output (or target) vectors $ \{ x_1 , \ldots , x_N \} \in \mathbb{R}^{M\times N}$.

Setting biases to zero, for the purpose of illustration, and using only one hidden layer ($L=2$) with $K < N $ factors, gives for $j=1, \ldots, N$
\begin{align*}
Y_j(x) = F^m_{W} ( X )_j & = \sum_{k=1}^K W^{jk}_2 f \left ( \sum_{i=1}^N W^{ki}_1 x_i \right )\\
& =  \sum_{k=1}^K W^{jk}_2 Z_j \,\, {\rm for }\,\, Z_j =  f \left ( \sum_{i=1}^N W^{ki}_1 x_i \right ).
\end{align*}

In an auto-encoder we fit the model $X = F_{W}( X) $, and \emph{train} the weights $ W = ( W_1 , W_2 ) $ with regularization penalty of the form
\begin{align*}
\mathcal{L} ( W )  =&\  {\rm arg \; min}_W \; \Vert X - F_W (X) \Vert^2  + \lambda \phi(W) \\
\text{with}\quad \phi(W) =&\  \sum_{i,j,k} | W^{jk}_1 |^2 +  | W^{ki}_2 |^2.
\end{align*}

Writing our DL objective as an augmented Lagrangian (as in ADMM) with a hidden factor $Z$, leads to a two step algorithm, an encoding step (a penalty for $Z$), and a decoding step for reconstructing the output signal via
$$
{\rm arg \; min}_{W,Z} \; \Vert X - W_2 Z \Vert^2 + \lambda \phi(Z) + \Vert Z -  f( W_1, X ) \Vert^2,
$$
where the regularization on $W_1$ induces a penalty on $Z$. The last term is the encoder, the first two the decoder.

If $ W_2$ is estimated from the structure of the training data matrix, then we have a traditional factor model, and the
$W_1$ matrix provides the factor loadings. PCA, PLS, SIR  fall into this category, see Cook (2007) for further discussion. 
If $W_2$ is trained based on the pair $\hat{X}=\{Y,X\}$ than we have a sliced inverse regression model. If $W_1$ and $W_2$ are simultaneously estimated based on the training data $X$, then we have a two layer deep learning model. 

Auto-encoding demonstrates that deep learning does not directly model variance-covariance matrix explicitly as the architecture is already in predictive form.
Given a hierarchical non-linear combination of deep learners, an implicit variance-covariance matrix exists, but that is not the focus of the algorithm. 

Another interesting area for future research are long short-term memory models (LSTMs).
For example, a dynamic one layer auto-encoder for a financial time series $(Y_t)$ is a coupled system
$$
Y_t = W_x X _t + W_y Y_{t-1} \; \; {\rm and} \; \; \left ( \begin{array}{c}
X_t\\
Y_{t-1} 
\end{array}
\right ) = W Y_t\,.
$$
The state equation encodes and 
the matrix $W$ decodes the $Y_t$ vector into its history $Y_{t-1}$ and the current state $X_t$.

\subsection{Bayesian Inference for Deep Learning}
Bayesian neural networks have a long history.  Early results on stochastic recurrent neural networks (a.k.a Boltzmann machines) were published in  \cite{ackley1985learning}. 
Accounting for uncertainty by integrating over parameters is discussed in \cite{denker1987large}. \cite{mackay1992practical} proposed a general Bayesian framework for tuning network architecture and training parameters for feed forward architectures.  \cite{neal1993bayesian}  proposed using Hamiltonian Monte Carlo (HMC) to sample from posterior distribution over the set of model parameters and then averaging outputs of multiple models. Markov Chain Monte Carlo algorithms was proposed by \cite{muller_issues_1998} to jointly identify parameters of a feed forward neural network as well as the architecture. A connection of neural networks with Bayesian nonparametric techniques was demonstrated in \cite{lee_bayesian_2004}. 

A Bayesian extension of feed forward network architectures has been considered by several authors \citep{neal1990learning,saul1996mean,frey1999variational,lawrence2005probabilistic,adams2010learning,mnih2014neural,kingma2013auto,rezende2014stochastic}. Recent results show how dropout regularization can be used to represent uncertainty in deep learning models.  In particular, \cite{Gal2015Theoretically} shows that dropout technique provides uncertainty estimates for the predicted values. The predictions generated by the deep learning models with dropout are nothing but samples from predictive posterior distribution.

Graphical models with deep learning encode a joint distribution via a product of conditional distributions and allow for computing (inference) many different probability distributions associated with the same set of variables. Inference requires the calculation of a posterior distribution over the variables of interest, given the relations between the variables encoded in a graph and the prior distributions. This approach is powerful when learning from samples with missing values or predicting with some missing inputs. 

A classical example of using neural  networks to model a vector of binary variables is the Boltzmann machine (BM), with two layers. The first layer encodes latent variables and the second layer encodes the observed variables. Both conditional distributions  $p(\text{data} \mid \text{latent variables})$ and $p(\text{latent variables} \mid \text{data})$ are specified using logistic function parametrized by weights and offset vectors. The size of the joint distribution table grows exponentially with the number of variables and \cite{hinton1983optimal} proposed using Gibbs sampler to calculate update to model weights on each iteration. The multimodal nature of the posterior distribution leads to prohibitive computational times required to learn models of a practical size. \cite{tieleman2008training} proposed a variational approach that replaces the  posterior $p(\text{latent variables} \mid \text{data})$  and approximates it with another easy to calculate distribution  was considered in \cite{salakhutdinov2008learning}. Several extensions to the BMs have been proposed. An exponential family extensions have been considered by \cite{smolensky1986,salakhutdinov2008learning,salakhutdinov2009deep,welling2005exponential}
 
There have also been multiple approaches to building inference algorithms for deep learning models \cite{mackay1992practical,hinton1993keeping,neal1992bayesian,barber1998ensemble}. Performing Bayesian inference on a neural network calculates the posterior distribution over the weights given the observations.  In general, such a posterior cannot be calculated analytically, or even efficiently sampled from. However, several recently proposed approaches address the computational problem for some specific deep learning models \citep{graves2011practical, kingma2013auto,rezende2014stochastic,blundell2015weight,hernandez2015probabilistic,gal2016dropout}.

%We need to include correlations when performing variational inference \cite{louizos2016structured}.

The recent successful approaches to develop efficient Bayesian inference algorithms for deep learning networks are based on the reparameterization techniques for calculating Monte Carlo gradients while performing  variational inference. Given the data $D = (X,Y)$, the variation inference relies on approximating the posterior $p(\theta \mid D)$ with a variation distribution  $q(\theta \mid D,\phi)$, where $\theta = (W,b)$. Then $q$ is found by minimizing the based on the Kullback-Leibler divergence between the approximate distribution and the posterior, namely
\[
\text{KL}(q \mid\mid p) = \int q(\theta \mid D, \phi)\log \dfrac{q(\theta \mid D, \phi)}{p(\theta\mid D)}d\theta.
\]
Since $p(\theta\mid D)$ is not necessarily tractable, we replace minimization of $\text{KL}(q \mid\mid p) $ with maximization of  evidence lower bound (ELBO)
\[
\text{ELBO}(\phi) = \int q(\theta \mid D,\phi)\log \dfrac{p(Y\mid X,\theta)p(\theta)}{q(\theta \mid D, \phi)}d\theta
\]
The $log$ of the total probability (evidence) is then
\[
\log p(D) =  \text{ELBO}(\phi) + \text{KL}(q \mid\mid p)
\]
The sum does not depend on $\phi$, thus minimizing $\text{KL}(q \mid\mid p)$ is the same that maximizing $\text{ELBO}(q) $. Also, since $\text{KL}(q \mid\mid p) \ge 0$, which follows from Jensen's inequality, we have $\log p(D) \ge  \text{ELBO}(\phi)$. Thus, the  evidence lower bound name.  The resulting maximization problem $\text{ELBO}(\phi) \rightarrow \max_{\phi}$ is solved using stochastic gradient descent. 

To calculate the gradient, it is convenient to write the ELBO as
\[
\text{ELBO}(\phi) = \int q(\theta \mid D, \phi)\log p(Y\mid X,\theta)d\theta - \int q(\theta \mid D,\phi) \log \dfrac{q(\theta\mid D, \phi)}{p(\theta)}d\theta
\]
The gradient of the first term $\nabla_{\phi}\int q(\theta \mid D, \phi)\log p(Y\mid X,\theta)d\theta = \nabla_{\phi}E_q\log p(Y\mid X,\theta)$ is not an expectation and thus cannot be calculated using Monte Carlo methods. The idea is to represent the gradient $\nabla_{\phi} E_q\log p(Y\mid X,\theta)$ as an expectation of some random variable, so that Monte Carlo techniques can be used to calculate it. There are two standard methods to do it. First, the log-derivative trick, uses the following identity $\nabla_x f(x) = f(x) \nabla_x \log f(x)$ to obtain $\nabla_{\phi} E_q\log p(Y\mid \theta)$.
%\[
%\nabla_{\phi} E_q\log p(Y\mid \theta) = \int q(\theta \mid Y, \phi)\nabla_{\phi}\log q(\theta \mid Y,\phi)\log p(Y\mid \theta)d\theta = E_q [\nabla_{\phi}\log q(\theta \mid Y,\phi)\log p(Y\mid \theta)].
%\]
Thus, if we select $q(\theta \mid\phi)$ so that it is easy to compute its derivative and generate samples from it, the gradient can be efficiently calculated using Monte Carlo technique.  Second, we can use reparametrization trick by representing $\theta$ as a value of a deterministic function,  $\theta = g(\epsilon,x,\phi)$, where $\epsilon \sim r(\epsilon)$ does not depend on $\phi$. The derivative is given by
\begin{align*}
\nabla_{\phi} E_q\log p(Y\mid X, \theta) &= \int r(\epsilon)\nabla_{\phi}\log p(Y\mid  X, g(\epsilon,x,\phi))d \epsilon\\
& = E_{\epsilon} [\nabla_{g}\log p(Y\mid X, g(\epsilon,x,\phi))\nabla_{\phi}g(\epsilon,x,\phi)].
\end{align*}
The reparametrization is trivial when $q(\theta \mid D,\phi) = N(\theta \mid \mu(D,\phi), \Sigma(D,\phi))$, and $\theta = \mu(D,\phi)  + \epsilon\Sigma(D,\phi),~\epsilon\sim N(0,I)$. \cite{kingma2013auto}  propose using $\Sigma(D,\phi) = I$ and representing $\mu(D,\phi)$ and $\epsilon$ as outputs of a neural network (multi-layer perceptron), the resulting approach was called variational auto-encoder. A generalized reparametrization has been proposed by \cite{ruiz2016generalized} and combines both log-derivative and reparametrization techniques by  assuming that $\epsilon$ can depend on $\phi$. 
%to obtain
%\[
%\nabla_{\phi} E_q\log p(Y\mid \theta) = \int r(\epsilon \mid \phi)\nabla_{\phi}\log p(Y\mid g(\epsilon,x,\phi))d e = E_{\epsilon} [\nabla_{g}\log p(Y\mid g(\epsilon,x,\phi))\nabla_{\phi}g(\epsilon,x,\phi)].
%\]

\section{Finding Good Bayes Predictors}
The Bayesian paradigm provides novel insights into how to construct estimators with good predictive performance. The goal is simply to find a good predictive MSE, namely $E_{Y,\hat{Y}}(\Vert\hat{Y} - Y \Vert^2)$, where $\hat{Y}$ denotes a prediction value. Stein shrinkage (a.k.a regularization with an $L^2$ norm) in known to provide good mean squared error properties in estimation, namely $E(||\hat{\theta} - \theta)||^2)$. These gains translate into predictive performance  (in an iid setting) for $E(||\hat{Y}-Y||^2)$. 

The main issue is how to tune the amount of regularisation (a.k.a prior hyper-parameters).  Stein's unbiased estimator of risk provides a simple empirical rule to address this problem as does cross-validation. From a Bayes perspective, the marginal likelihood (and full marginal posterior) provides a natural method for hyper-parameter tuning. The issue is computational tractability and scalability. In the context of DL, the posterior for $(W,b)$ is  extremely high dimensional and multimodal and posterior MAP provides good predictors $\hat{Y}(X)$. 

\begin{figure}[H]
	\centering
\begin{tabular}{cc}
	\includegraphics[width=0.5\textwidth]{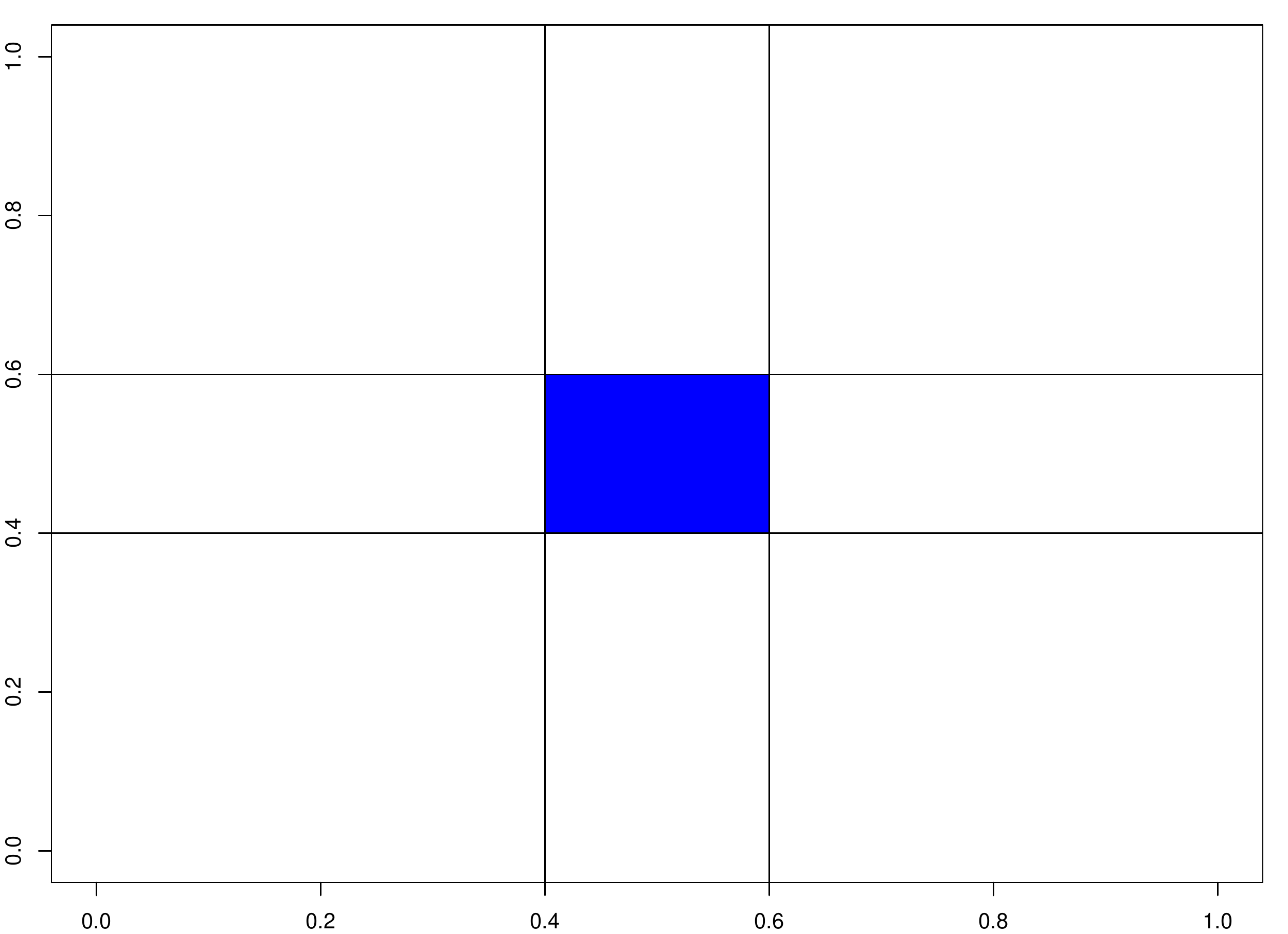}& 	\includegraphics[width=0.5\textwidth]{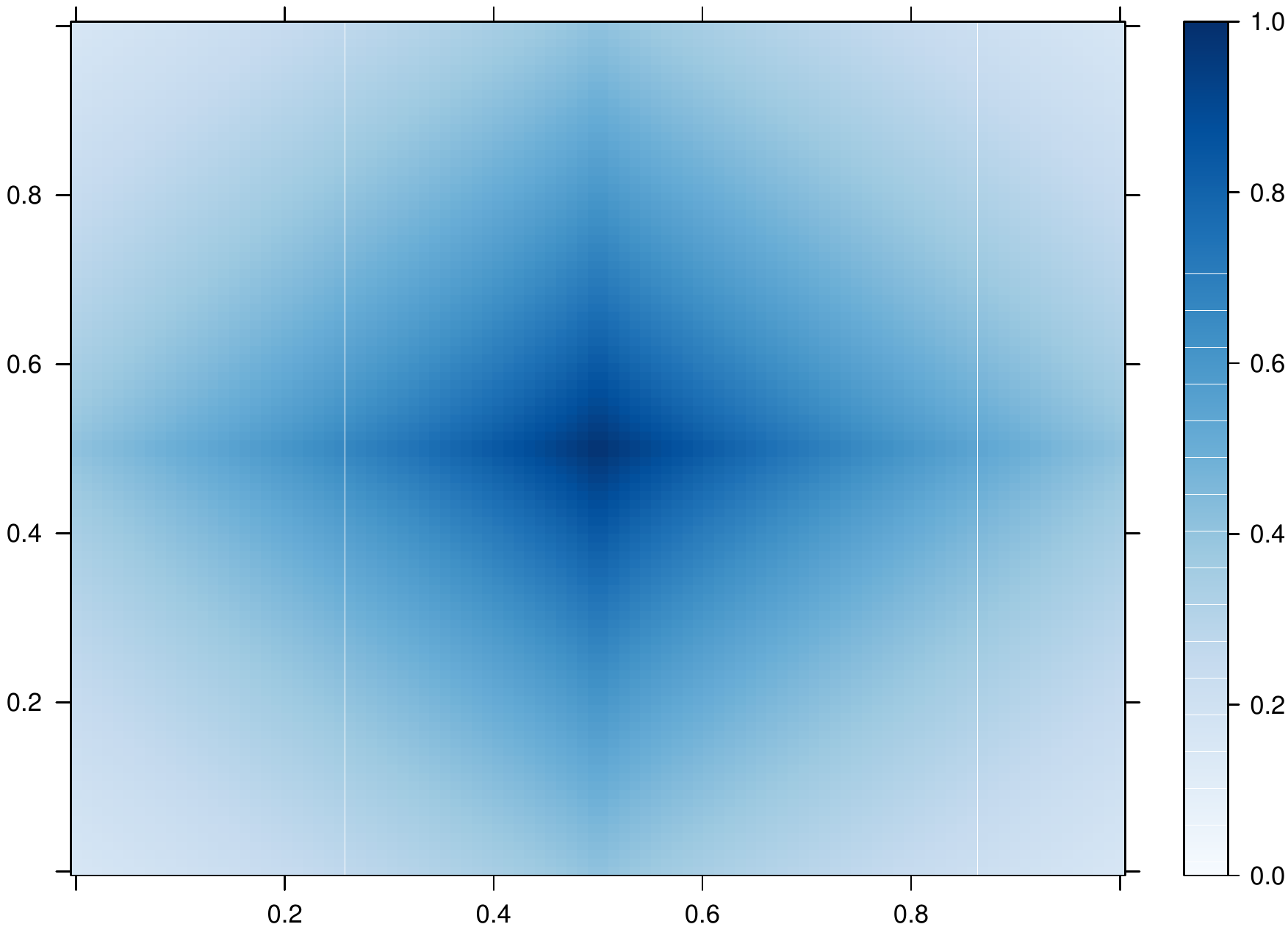}\\
	(a) Tree Kernel & (b) Random Forest Kernel
\end{tabular}
\caption{Kernel Weight. The intensity of the color is proportional to the size of the weight. Left panel (a) shows weights for tree-based model, with non-zero values only inside a cylindrical region (a box), and (b)  shows weights for a random forest model, with non-zero wights everywhere in the domain and sizes decaying away from the location of the new observation.}
\label{fig:cilinder}
\end{figure}

Bayes conditional averaging performs well in high dimensional regression and classification problems. High dimensionality, however, brings with it the curse of dimensionality and  it is instructive to understand why certain kernel can perform badly. Adaptive Kernel predictors (a.k.a. smart conditional averager) are of the form 
$$
\hat{Y}(X) = \sum_{r=1}^R K_r ( X_i , X ) \hat{Y}_r (X).
$$
Here $ \hat{Y}_r(X) $ is a deep predictor with its own trained parameters. For tree models, the kernel $ K_r( X_i , X) $ is a \emph{cylindrical} region $ R_r $ (open box set). Figure \ref{fig:cilinder} illustrates the implied kernels for trees (cylindrical sets) and random forests. Not too many points will be neighbors in a high dimensional input space.  

Constructing the regions to preform conditional averaging is fundamental to reduce the curse of dimensionality. Imagine a large dataset, e.g. 100k images and think about how a new image's input coordinates, $X$, are ``neighbors" to data points in the training set.  Our predictor is a smart conditional average of the observed outputs, $Y$, from our neighbors. When $p$ is large,  spheres ($L^2$ balls or Gaussian kernels) are terrible, degenerate cases occur when  either no points or all of the points are ``neighbors" of the new input variable will appear. Tree-based models address this issue by limiting the number of ``neighbors. 

\begin{figure}[H]
\centering
\includegraphics[width=0.5\textwidth]{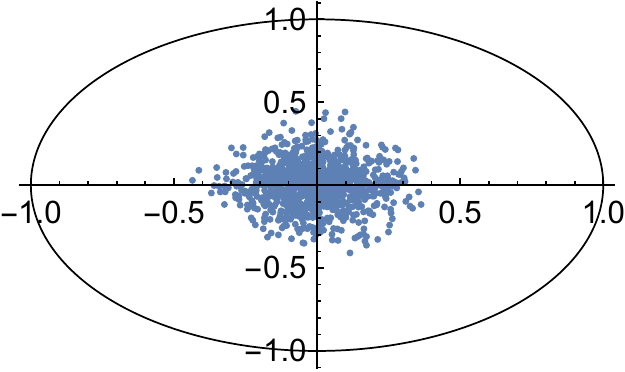}
\caption{The 2D image of the Monte Carlo samples from a the 50-dimensional ball.}
\label{fig:hd_ball}
\end{figure}

Figure \ref{fig:hd_ball} further illustrates the challenge with the 2D image of 1000 uniform samples from a 50-dimensional ball $B_{50}$. The image is calculated as $w^T Y$, where $w = (1,1,0,\ldots,0)$ and $Y \sim U(B_{50})$. Samples are centered around the equators and none of the samples fall anywhere close to the boundary of the set. 

\begin{figure}[H]
\begin{tabular}{cccc}
\includegraphics[width=0.25\textwidth]{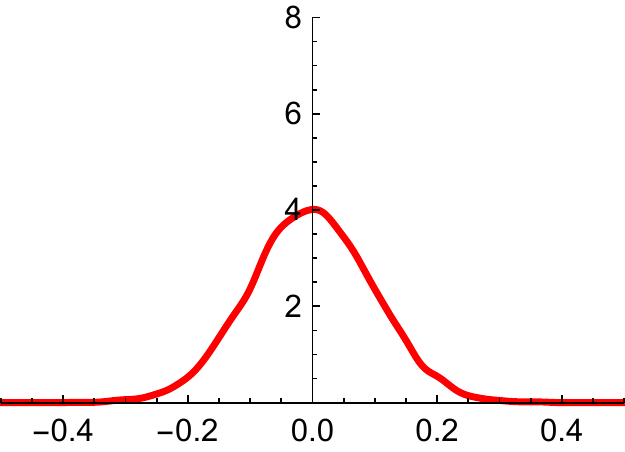}&
\includegraphics[width=0.25\textwidth]{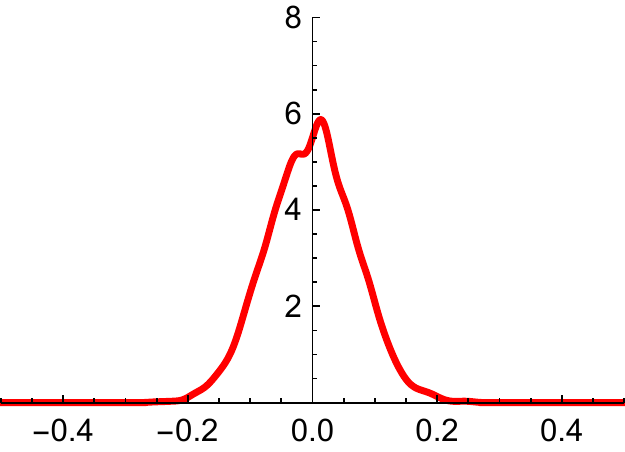}&
\includegraphics[width=0.25\textwidth]{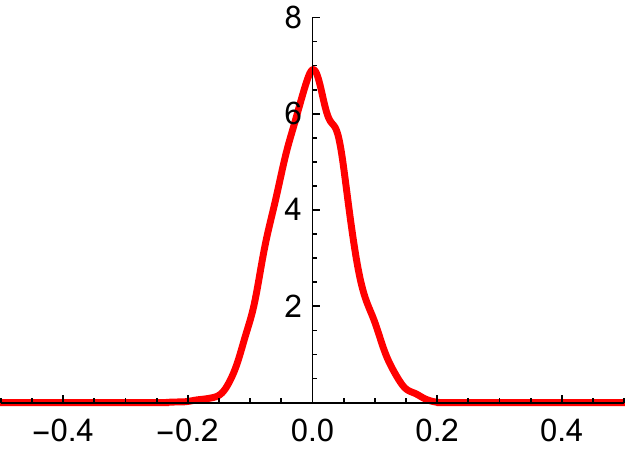}&
\includegraphics[width=0.25\textwidth]{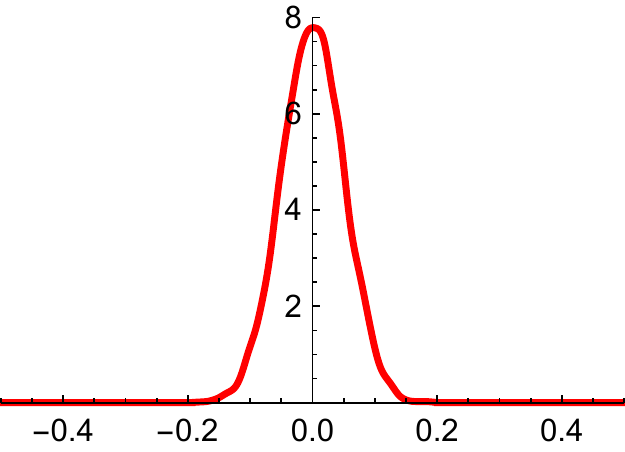}\\
(a) $p$ = 100 & (b) $p$ = 200 &(c) $p$ = 300 &(d) $p$ = 400
\end{tabular}
\caption{Histogram of marginal distributions of $Y\sim U(B_p)$ samples for different dimensions $p$.}
\label{fig:hd_hist}
\end{figure}

As dimensionality of the space grows, the variance of the marginal distribution goes to zero. Figure \ref{fig:hd_hist} shows the histogram of 1D image of uniform sample from balls of different dimensionality, that is $e_1^T Y$, where $e_1 = (1,0,\ldots,0)$.

Similar central limit results were known to Maxwell  who has shown that the random variable $w^TY$ is close to standard normal, when $Y \sim U(B_p)$, $p$ is large, and $w$  is a unit vector (lies on the boundary of the ball), see \cite{diaconis1987dozen}. More general results in this direction were obtained in \cite{klartag2007central} and \cite{milman2009asymptotic} who presents many analytical and geometrical  results for finite dimensional normed spaces, as the dimension grows to infinity.

Deep learning can improve on traditional methods by performing a sequence of GLM-like transformations. Effectively DL learns a distributed partition of the input space. 
For example, suppose that we have $K$ partitions and a DL predictor that takes the form of a weighted average or soft-max of the weighted average for classification.  Given a new high dimensional input $X_{\mathrm{new}}$, many deep learners are then an average of learners obtained by our hyper-plane decomposition. Our predictor takes the form
\[
\hat{Y}(X) = \sum_{k \in K} w_k(X)\hat{Y}_k(X),
\] 
where $w_k$ are the weights learned in region $K$, and $k$ is an indicator of the region with appropriate weighting given the training data.

\begin{figure}[H]
	\centering
\includegraphics[width=0.5\textwidth]{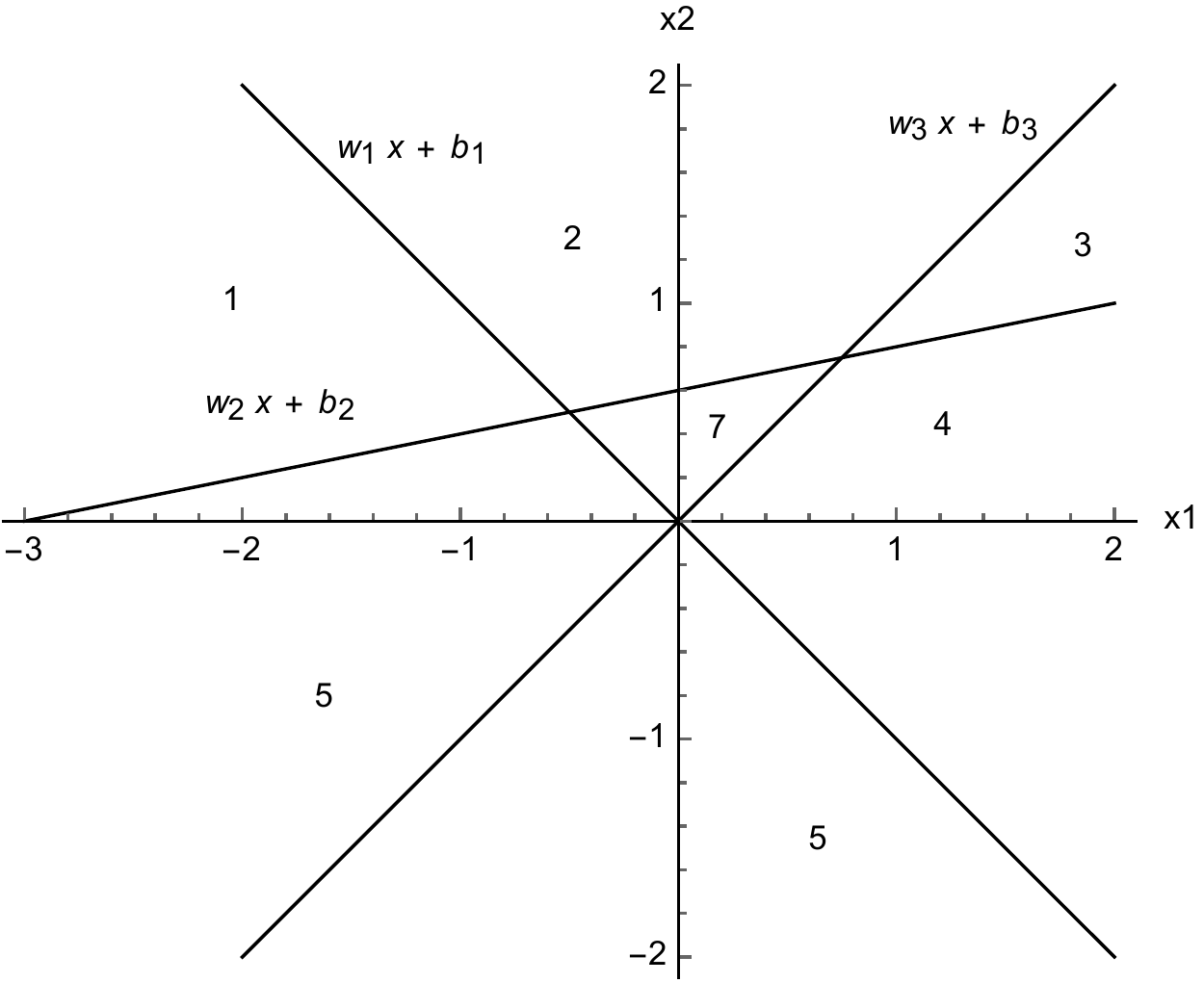}
\caption{Hyperplanes defined by three neurons with ReLU activation functions.}
\label{fig:hyper_planes}
\end{figure}

The partitioning of the input space by a deep learner is similar to the
one performed by decision trees and partition-based models such as
CART, MARS, RandomForests, BART, and Gaussian Processes. Each neuron in
a deep learning model corresponds to a manifold that divides the input
space. In the case of ReLU activation function $f(x) = \max(x,0)$ the
manifold is simply a hyperplane and the neuron gets activated when the new
observation is on the ``right'' side of this hyperplane, the activation
amount is equal to how far from the boundary the given point is. For
example in two dimensions, three neurons with ReLU activation functions
will divide the space into seven regions, as shown on Figure \ref
{fig:hyper_planes}.

\begin{figure}[H]
	\begin{tabular}{ccc}
		\includegraphics[width=0.33\textwidth]{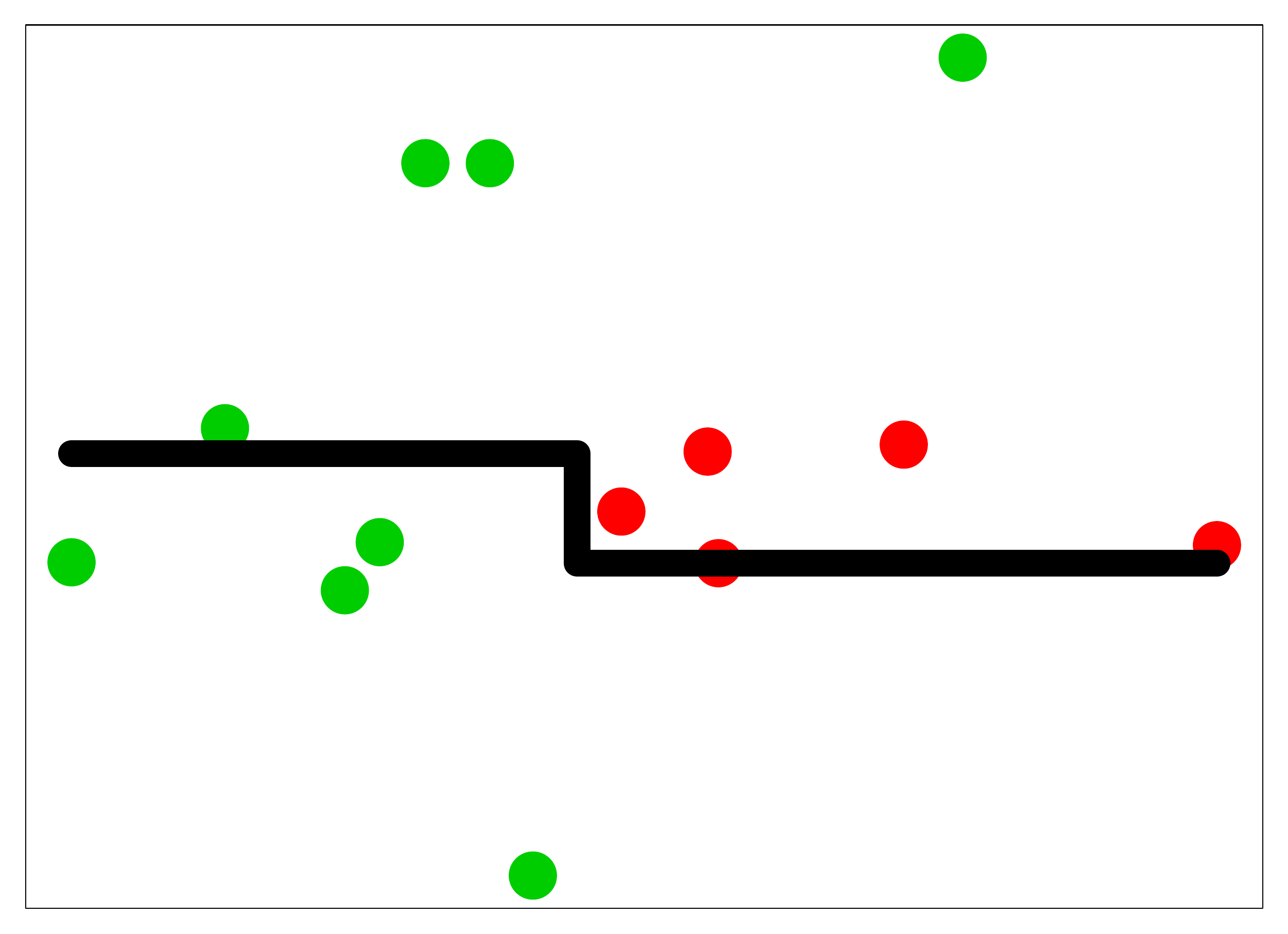} &
		\includegraphics[width=0.33\textwidth]{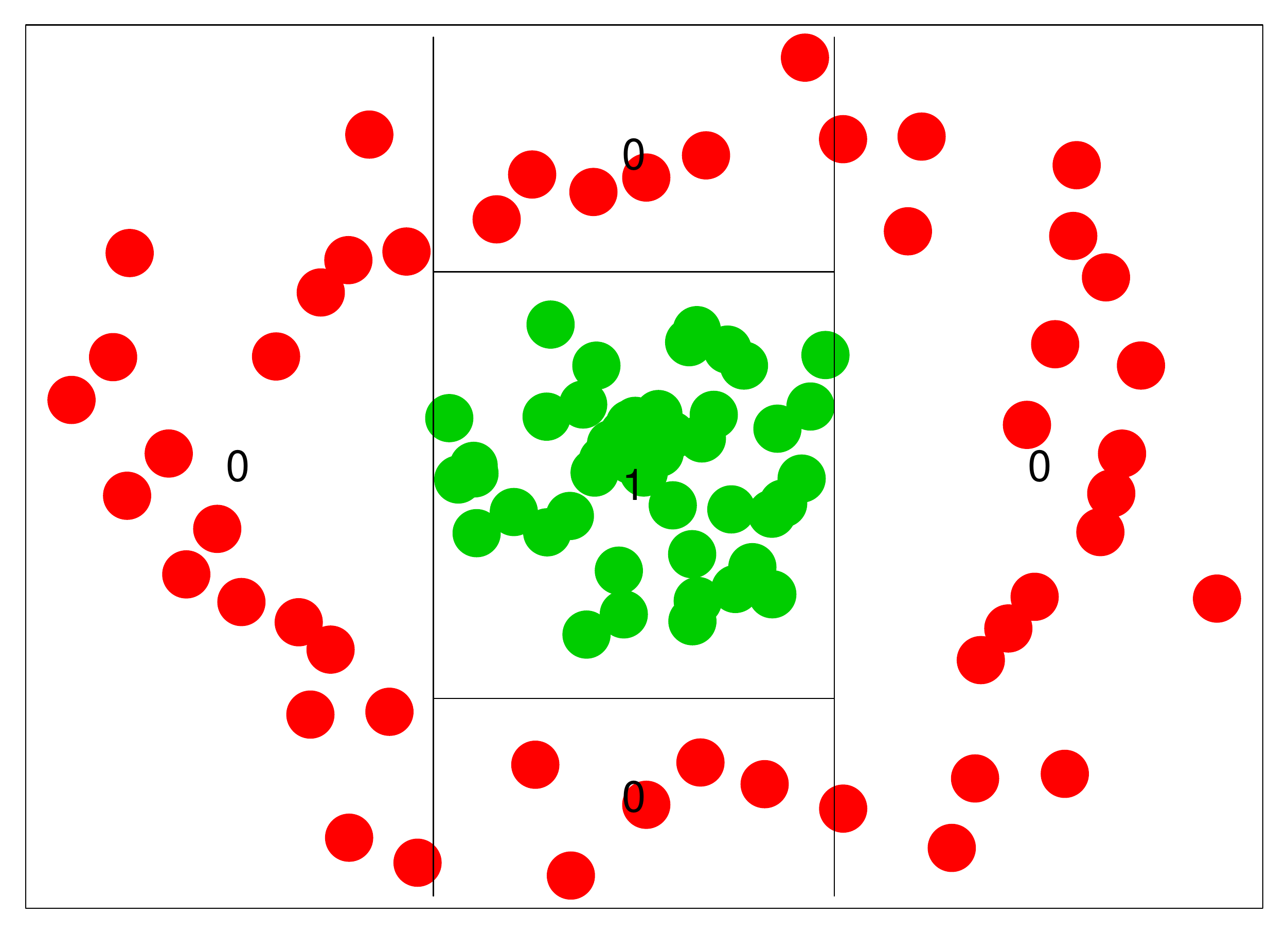} &
		\includegraphics[width=0.33\textwidth]{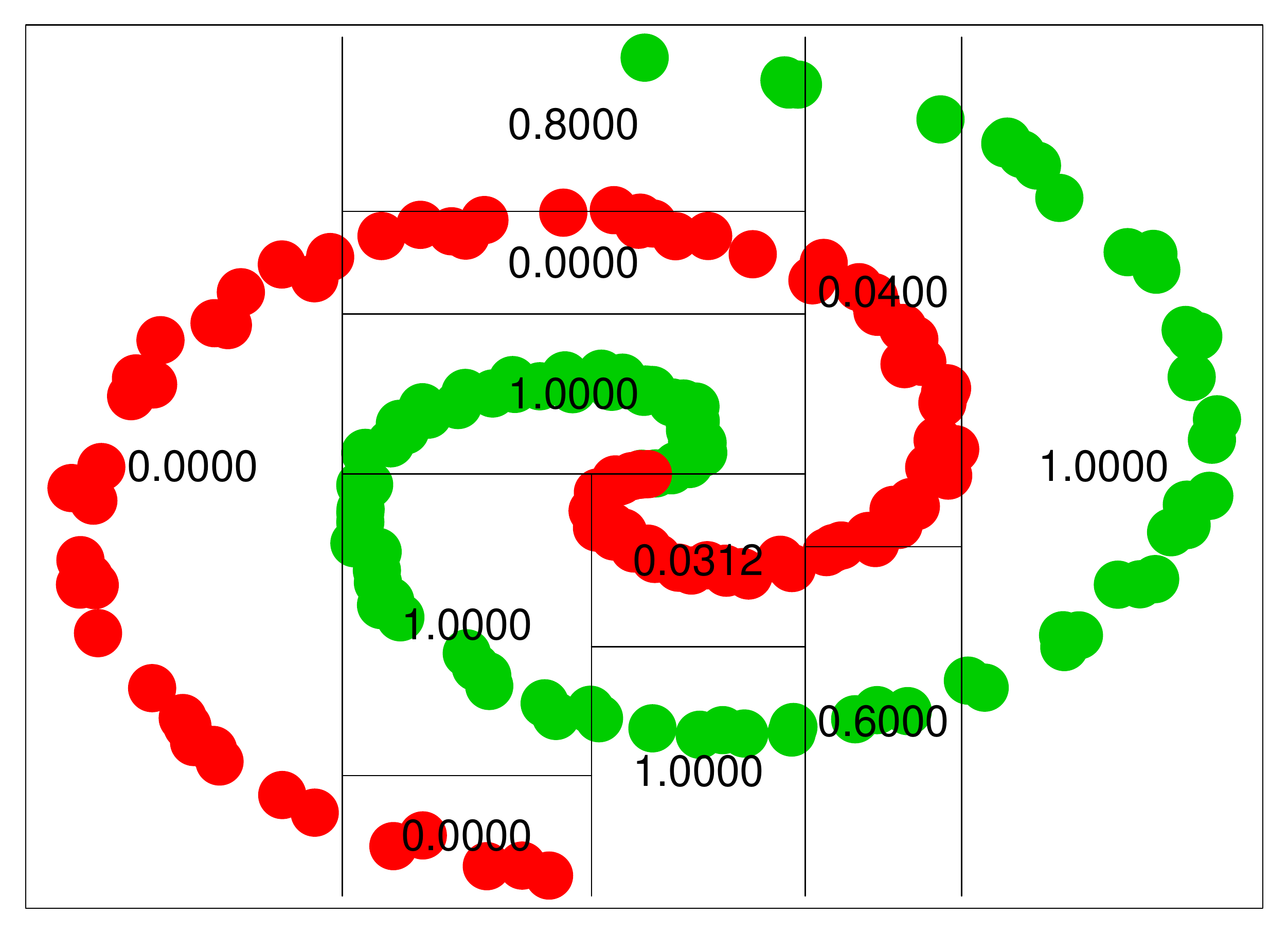} \\
		\includegraphics[width=0.33\textwidth]{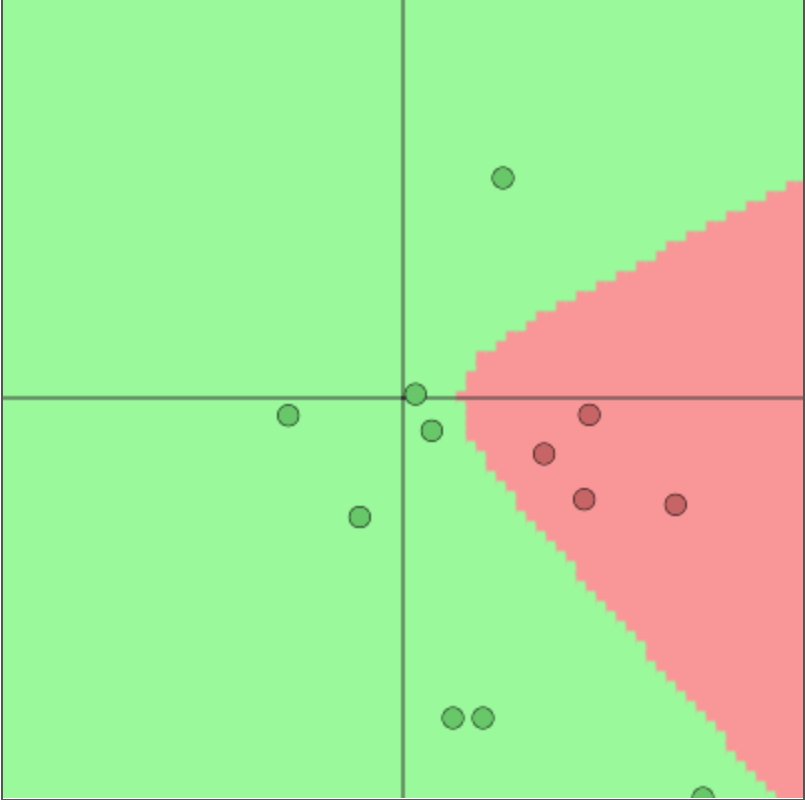} &
		\includegraphics[width=0.33\textwidth]{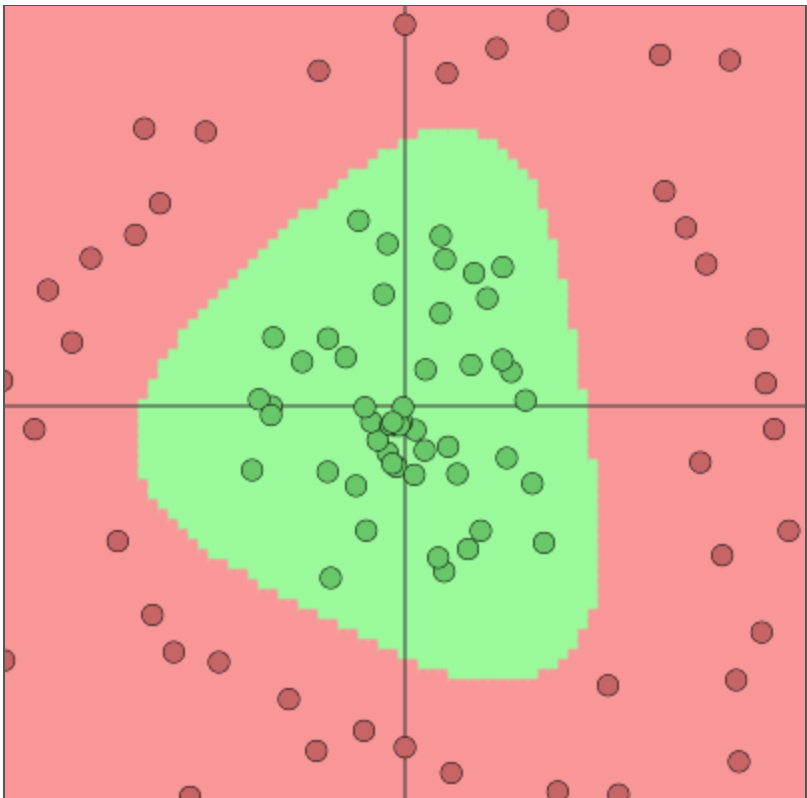} &
		\includegraphics[width=0.33\textwidth]{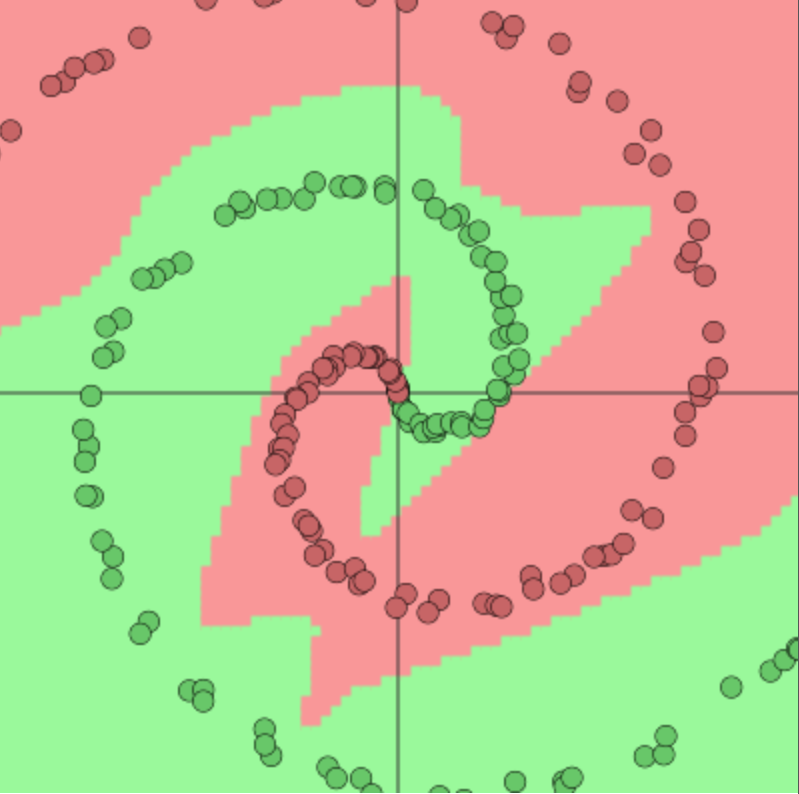} \\
		(a) simple data & (b) circle data & (c) spiral data
	\end{tabular}
	\caption{Space partition by tree architectures (top row) and deep learning architectures (bottom row) for three different data sets.}
	\label{fig:tree-dl-comp}
\end{figure}

The key difference between tree-based architecture and neural
network based models is the way hyper-planes are combined. Figure \ref
{fig:tree-dl-comp} shows the comparison of space decomposition by
hyperplanes, as performed by a tree-based and neural network
architectures. We compare a neural network with two layers (bottom row)
with tree mode trained with CART algorithm (top row). The network
architecture is given by
\begin{align*}
Y =  &\mathrm{softmax}(w^0Z^2 + b^0)\\
Z^2 = & \tanh(w^2Z^1 + b^2)\\
Z^1 = & \tanh(w^1X + b^1) \; .
\end{align*}
The weight matrices for simple data $W^1, W^2 \in \mathbb{R}^{2 \times 2}$, for circle data $W^1 \in \mathbb{R}^{2 \times 2}$ and $W^2 \in \mathbb{R}^{3 \times 2}$, for spiral data we have $W^1 \in \mathbb{R}^{2 \times 2}$ and $W^2 \in \mathbb{R}^{4 \times 2}$. In our notations, we assume that the activation function is applied point-vise at each layer. 
An advantage of deep architectures is that the number of hyper-planes grow exponentially with the number of layers. The key property of an activation function (link) is $f(0) = 0$ and it has zero value in certain regions. For example, hinge or rectified learner $\max(x,0)$  box car (differences in Heaviside) functions are very common. As compared to  a logistic regression, rather than using $\mathrm{softmax}(1/(1+e^{-x}))$ in deep learning $\tanh(x)$ is typically used for training, as $\tanh(0)=0$. 

%\begin{figure}[H]
%	\includegraphics[width=1\textwidth]{tree-dl-comp}
%	\caption{Comparison of space decomposition by hyperplanes.}
%	\label{fig:tree-dl-comp}
%\end{figure}

\cite{amit_shape_1997} provide an interesting discussion of efficiency. Formally, a Bayesian probabilistic approach (if computationally feasible) optimally weights predictors via model averaging with  $\hat{Y}_k(x) = E(Y \mid X_k)$
$$
\hat{Y}(X) = \sum_{r=1}^R w_k \hat{Y}_k(X).
$$

Such rules can achieve optimal out-of-sample performance.  \cite{amit2000multiple} discusses the striking success of multiple randomized classifiers. Using a simple set of binary local features, one classification tree can achieve 5\% error on the NIST data base with 100,000 training data points. On the other hand, 100 trees, trained under one hour, when aggregated, yield an error rate under 7\%. This stems from the fact that a sample from a very rich and diverse set of classifiers produces, on average, weakly dependent classifiers conditional on class. 

To further exploit this, consider the Bayesian model of weak dependence, namely exchangeability. Suppose that we have  $K$ exchangeable, $ \mathbb{E} ( \hat{Y}_i ) = \mathbb{E} ( \hat{Y}_{\pi(i)} ) $, and stacked predictors
$$
\hat{Y} = ( \hat{Y}_1 , \ldots , \hat{Y}_K ). 
$$
Suppose that we wish to find weights, $w$, to attain $ {\rm arg \; min}_W \; E l( Y , w^T \hat{Y} ) $ where $l$ convex in the second argument;
\[
E l( Y , w^T \hat{Y} )  = \frac{1}{K!} \sum_\pi E l( Y , w^T \hat{Y} )
 \geq  E l \left ( Y , \frac{1}{K!} \sum_\pi w_\pi^T \hat{Y} )\right ) =  E l \left ( Y , (1/K) \iota^T \hat{Y} \right ) 
\]
where $ \iota = ( 1 , \ldots ,1 ) $. Hence, the randomised multiple predictor with weights $w = (1/K)\iota$  provides the optimal Bayes predictive performance. 

We now turn to algorithmic issues. 

\section{Algorithmic Issues}\label{sec:algorithms}
In this section we discuss two types of algorithms for training learning models. First, stochastic gradient descent, which is a very general algorithm that efficiently works for large scale datasets and has been used for many deep learning applications. Second, we discuss specialized statistical learning algorithms, which are tailored for certain types of traditional statistical models.  

\subsection{Stochastic Gradient Descent}
Stochastic gradient descent (SGD) is a default gold standard for minimizing
the a function $f(W,b)$ (maximizing the likelihood) to find the deep
learning weights and offsets. SGD simply minimizes the function by
taking a negative step along an estimate $g^k$ of the gradient $\nabla
f(W^k, b^k)$ at iteration $k$. The gradients are available via the
chain rule applied to the superposition of semi-affine functions. 

The
approximate gradient is estimated by calculating
\[
g^k = \frac{1}{|E_k|} \sum_{i \in E_k} \nabla\mathcal{L}_{w,b}( Y_i ,
\hat{Y}^k( X_i)),
\]
where $E_k \subset\{1,\ldots,T \}$ and $|E_k|$ is the number of
elements in $E_k$.\vadjust{\goodbreak}

When $|E_k| >1$ the algorithm is called batch SGD and simply SGD
otherwise. Typically, the subset $E$ is chosen by going cyclically and
picking consecutive elements of $\{1,\ldots,T \}$, $E_{k+1} = [E_k \mod
T]+1$. The direction $g^k$ is calculated using a chain rule (a.k.a.
back-propagation) providing an unbiased estimator of $\nabla f(W^k,
b^k)$. Specifically, this leads to
\[
\mathrm{E}(g^k) = \frac{1}{T} \sum_{i =1}^T \nabla\mathcal{L}_{w,b}(
Y_i , \hat{Y}^k( X_i)) = \nabla f(W^k, b^k).
\]
At each iteration,  SGD updates the solution
\[
(W,b)^{k+1} = (W,b)^k - t_k g^k.
\]
Deep learning algorithms use a step size $t_k$ (a.k.a learning rate) that is either kept constant or a simple step size reduction strategy, such as $t_k = a\exp(-kt)$  is used. The hyper parameters of reduction schedule  are usually found empirically from numerical experiments and observations of the loss function progression. 

One caveat of SGD is that the descent in $f$ is not guaranteed, or it can be very slow at every iteration. Stochastic Bayesian approaches ought to alleviate these issues.  The variance of the gradient estimate $g^k$ can also be  near zero, as the iterates converge to a solution.  To tackle those problems a coordinate descent (CD) and momentum-based modifications  can be applied. Alternative directions method of multipliers (ADMM) can also provide a natural alternative, and leads to non-linear alternating updates, see \cite{carreira2014distributed}. 

The  CD evaluates a single component $E_k$ of the gradient $\nabla f$ at the current point and then updates the $E_k$th component of the variable vector in the negative gradient direction. The momentum-based versions of SGD, or so-called accelerated algorithms were originally proposed by  \cite{nesterov1983method}. For more recent discussion, see \cite{nesterov2013introductory}. 
The momentum term adds memory to the search process by combining  new gradient information with the previous search directions. Empirically momentum-based methods have been shown a better convergence for deep learning networks \cite{sutskever2013importance}. The gradient only influences changes in the velocity of the update, which then updates the variable
\begin{align*}
v^{k+1} =     & \mu v^k - t_k g((W,b)^k)\\
(W,b)^{k+1} = & (W,b)^k +v^k
\end{align*}
The hyper-parameter $\mu$ controls the dumping effect on the rate of update of the variables. The physical analogy is the reduction in kinetic energy that allows to ``slow down" the movements at the minima. This parameter can also be chosen empirically using cross-validation. 

Nesterov's momentum method (a.k.a. Nesterov acceleration) calculates the gradient at the point predicted by the momentum. One can view this as a look-ahead strategy with updating scheme
\begin{align*}
v^{k+1} = & \mu v^k - t_k g((W,b)^k +v^k)\\
(W,b)^{k+1} = & (W,b)^k +v^k.
\end{align*}
Another popular modification are the AdaGrad methods \cite{zeiler2012adadelta}, which adaptively scales each of the learning parameter at each iteration 
\begin{align*}
c^{k+1} = &      c^k + g((W,b)^k)^2\\
(W,b)^{k+1} = & (W,b)^k - t_k g(W,b)^k)/(\sqrt{c^{k+1}} - a),
\end{align*}
where is usually a small number, e.g. $a = 10^{-6}$ that prevents dividing by zero. {\tt PRMSprop} takes the {\tt AdaGrad} idea further and places more weight on recent values of gradient squared to scale the update direction, i.e. we have 
\[
c^{k+1} =  dc^k + (1-d)g((W,b)^k)^2\\
\]
The {\tt Adam} method \citep{kingma2014adam} combines both {\tt PRMSprop} and momentum methods, and leads to the following update equations
\begin{align*}
v^{k+1} = & \mu v^k - (1-\mu)t_k g((W,b)^k +v^k)\\
c^{k+1} = & dc^k + (1-d)g((W,b)^k)^2\\
(W,b)^{k+1} = & (W,b)^k - t_k v^{k+1}/(\sqrt{c^{k+1}} - a) 
\end{align*}
Second order methods solve the optimization problem by solving a system of nonlinear equations $\nabla f(W,b) = 0$ by applying the Newton's method
\[
(W,b)^+ = (W,b) - \{ \nabla^2f(W,b) \}^{-1}\nabla f(W,b).
\]
Here SGD simply approximates $\nabla^2f(W,b)$ by $1/t$. The advantages of a second order method include much faster convergence rates and insensitivity to the conditioning of the problem. In practice, second order methods are rarely used for deep learning applications \citep{dean2012}. The major disadvantage is its inability to train models using batches of data as SGD does. Since a typical deep learning model relies on large scale data sets, second order methods become memory and computationally prohibitive at even modest-sized training data sets.

\subsection{Learning Shallow Predictors}
Traditional factor models use linear combination of $K$ latent factors, $\{ z_1 , z_2 , \ldots , z_K \} $, 
\[
Y_i = \sum_{k=1}^K w_{ik} z_k \; , \; \forall i = 1 , \ldots , N.
\]
Here factors $z_k$ and weights $B_{ik}$ can be found by solving the following problem
\[
{\rm arg min}_{ w , F} \; \sum_{n=1}^N \Vert z_n - \sum_{k=1}^K w_{nk} z_k \Vert^2 + \lambda \sum_{k,n=1}^{N,K} \Vert w_{nk} \Vert_l.
\]
Then, we minimize the reconstruction error (a.k.a. accuracy), plus the regularization penalty, to control the variance-bias trade-off for out-of-sample prediction. Algorithms exist to solve this problem very efficiently. Such a model can be represented as a neural network model with $L=2$ with identity activation function. 

The basic sliced inverse regression (SIR) model takes the form  $Y = G(WX,\epsilon)$, where $G(\cdot)$ is a nonlinear function and $W \in R^{k \times p}$, with $k <p$, in other words, $Y$ is a function of $k$ linear combinations of $X$. To find $W$, we first slice the feature matrix, then we analyze the data's covariance matrices and slice means of $X$, weighted by the size of slice. The function $G$ is found empirically by visually exploring relations. The key advantage of deep learning approach is that functional relation $G$ is found automatically. To extend the original SIR fitting algorithm, \cite{jiang2013sliced} proposed a variable selection under the SIR modeling framework.
A partial least squares regression (PLS) \citep{wold2001} finds $T$, a lower dimensional representation of   $X = TP^T$ and then regresses it onto $Y$ via  $Y = TBC^T$.

A deep learning least squares network arrives at a criterion function given by  a negative log-posterior, which needs to be minimized. The penalized log-posterior, with $ \phi $ denoting a generic regularization penalty is given by
\begin{align*}
L( G , F ) = &\sum_{i=1}^n \enorm{ y_i - g (F (x_i) ) } + \lambda_g \phi (G,F)\\
\phi_i = &\sum_k f_k^L(a_{ik}^L),~a_{ik}^L = z_{ik}^Lw^L,~z_{ik}^L = \sum_j f^{L-1}(a^{L-1}_{jk}).
\end{align*}
\cite{carreira2014distributed} propose a method of auxiliary coordinates which replaces the original unconstrained optimization problem, associated with model training, with an alternative function in a constrained space, that can be optimized using alternating directions method and thus is highly parallelizable. 
An extension of these methods are ADMM and Divide and Concur (DC) algorithms, for further discussion see \cite{polson_proximal_2015}.
The gains for applying these to deep layered models, in an iterative fashion, appear to be large but have yet to be quantified empirically.

\section{Application: Predicting Airbnb Bookings}
To illustrate our methodology, we use the dataset provided by the
Airbnb Kaggle competition. This dataset whilst not designed to optimize
the performance of DL provides a useful benchmark to compare and
contrast traditional statistical models. The goal is to build a
model that can predict  which country a new user will make his or her
first booking. Though Airbnb offers bookings in more than 190
countries, there are 10 countries where users make frequent bookings.
We treat the problem as classification into one of the 12 classes (10
major countries + other + NDF); where \textit{other} corresponds to any
other country which is not in the list of top 10 and \textit{NDF}
corresponds to situations where no booking was made.

\begin{figure}[H]
\begin{tabular}{ccc}
	\includegraphics[width=0.3\textwidth]{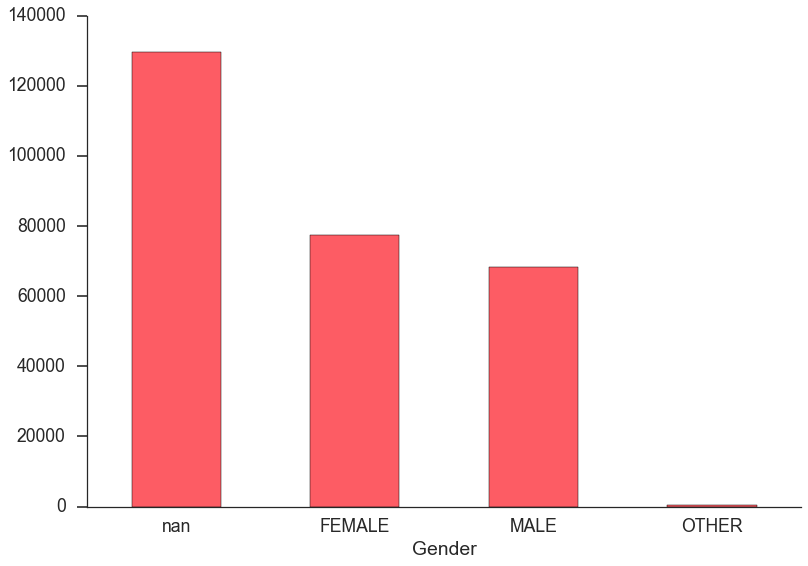} & \includegraphics[width=0.3\textwidth]{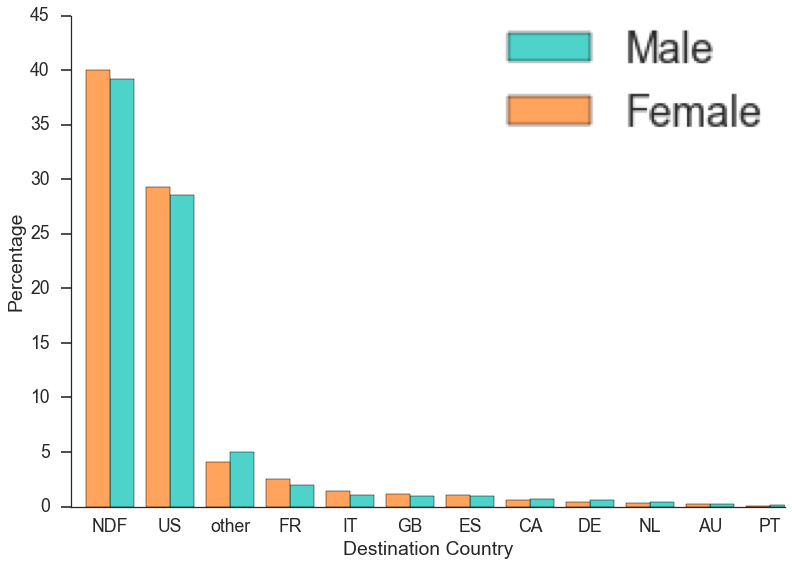} & \includegraphics[width=0.3\textwidth]{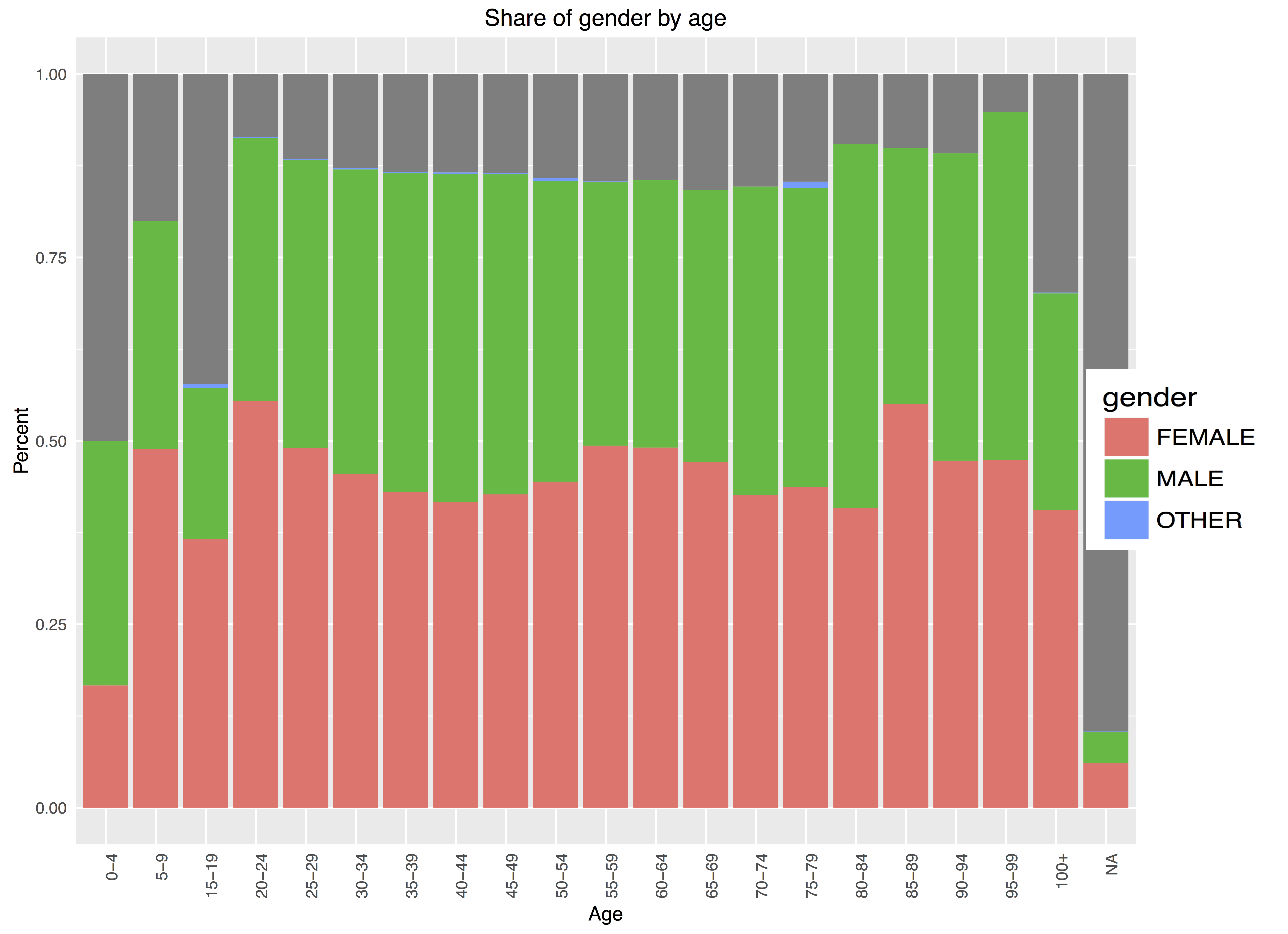}\\
	(a) Number of observations & (b) Percent of reservations & (c) Relationship between \\
	for each gender & per destination & age and gender
\end{tabular}
\caption{Gender and Destination Summary Plots for Airbnb users.}
\label{fig:airbnb_gender}
\end{figure}

The data consists of two tables, one contains the attributes of each of the users and the other contains data about sessions of each user at the Airbnb website. The user data contains demographic characteristics, type of device and browser used to sign up, and the destination country of the first booking, which is our dependent variable $Y$. The data involves 213,451 users  and 1,056,7737 individual sessions. The sessions data contains  information about actions taken during each session, duration and devices used. Both datasets has a large number of missing values. For example age information is missing for 42\% of the users. Figure \ref{fig:airbnb_gender}(a) shows that nearly half of the gender data is missing and there is slight imbalance between the genders. 

Figure \ref{fig:airbnb_gender}(b) shows the country of origin for the first booking by gender. Most of the entries in the destination columns are NDF, meaning no booking was made by the user. Further, Figure \ref{fig:airbnb_gender}(c) shows relationship between gender and age, the gender value is missing for most of the users who did not identify their age. 

We find that there is little difference in booking behavior between the genders. However, as we will see later, the fact that gender was specified, is an important predictor. Intuitively, users who filled the gender field are more likely to book. 

\begin{figure}[H]
	\begin{tabular}{ccc}
		\includegraphics[width=0.33\textwidth]{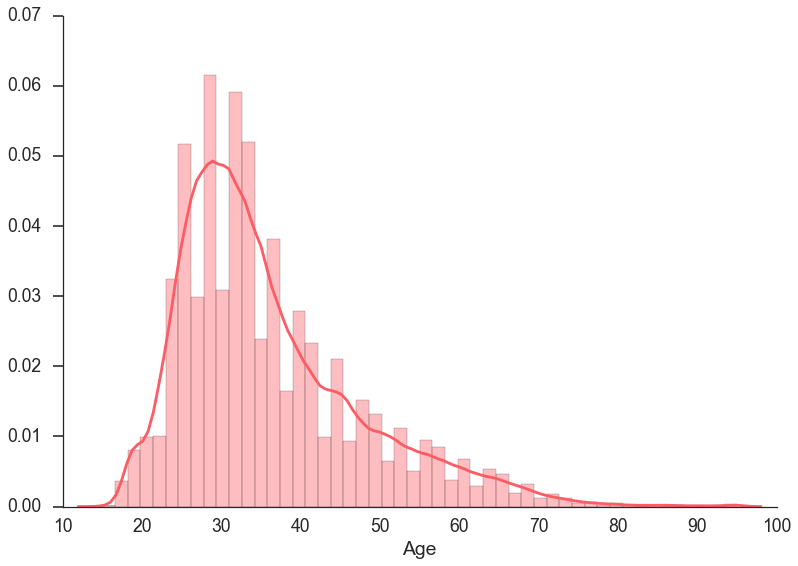} & \includegraphics[width=0.33\textwidth]{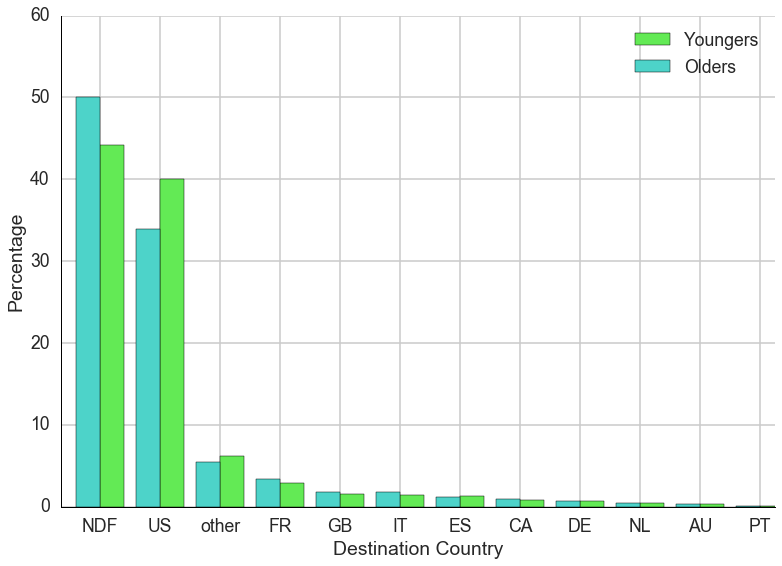} & \includegraphics[width=0.33\textwidth]{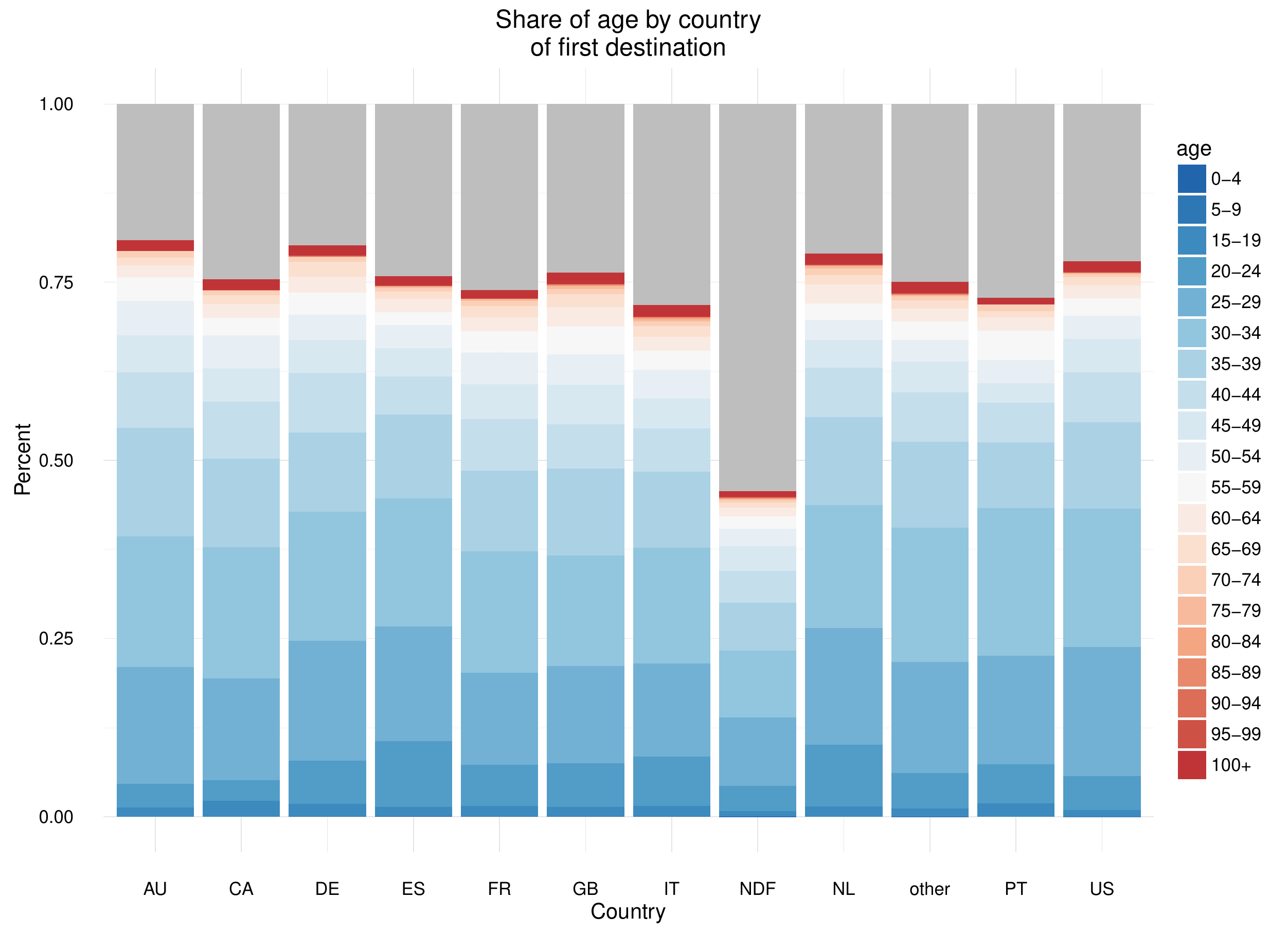}\\
		(a) Empirical distribution of & (b) Destination by age & (c) Destination by age \\
		user's age & category & group
	\end{tabular}
	\caption{Age Information for Airbnb users. }
	\label{fig:airbnb_age}
\end{figure}

On the other hand, as Figure \ref{fig:airbnb_age} shows, the age variable does play a role.

Figure \ref{fig:airbnb_age}(a) shows that most of the users are of age between 25 and 40. Furthermore, looking at booking behavior between two different age groups, younger than 45 cohort and older than 45 cohort, (see Figure \ref{fig:airbnb_age}(b)) have very different booking behavior. Further, as we can see from Figure \ref{fig:airbnb_age}(c) half of the users who did not book did not identify their age either.

\begin{figure}
%\vspace{-30pt}
\begin{center}
	\includegraphics[width=0.48\textwidth]{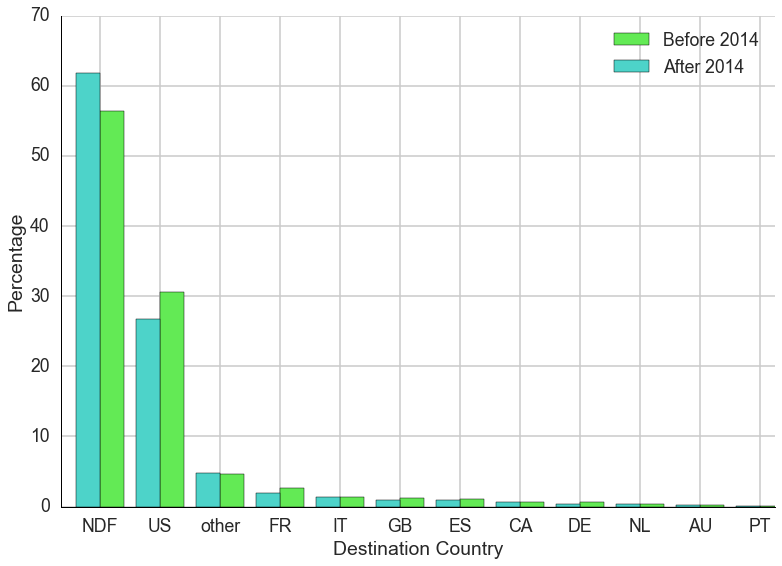}
\end{center}
%\vspace{-30pt}
\caption{Booking behavior for users who opened their accounts before 2014}
%\vspace{-30pt}
\label{fig:account_active}
\end{figure}
Another effect of interest is the non-linearity between the  time the account was created and booking behavior. Figure \ref{fig:account_active} shows that ``old timers" are more likely to book when compared to recent users. Since the number of records in sessions data is different for each users, we developed features from those records so that sessions data can be used for prediction. The general idea is to convert multiple session records to a single set of features per user. The list of the features we calculate is 
\begin{enumerate}[label=(\roman*)] 
	\item Number of sessions records
	\item For each action type, we calculate the count and standard deviation
	\item For each device type, we calculate the count and standard deviation
	\item For session duration we calculate mean, standard deviation and median
\end{enumerate}
Furthermore, we use one-hot encoding for categorical variables from the user table, e.g. gender, language, affiliate provider, etc. One-hot encoding replaces categorical variable with $K$ categories by $K$ binary dummy variable.

We build a deep learning model with two hidden dense layers and ReLU activation function $f(x) = \max(x,0)$. We use ADAGRAD optimization  to train the model. We predict probabilities of future destination booking for each of the new users. The evaluation metric for this competition is NDCG (Normalized discounted cumulative gain). We use top five predicted destinations and is calculated as:

\[ \mathrm{NDCG}_k= \dfrac{1}{n}\sum_{i=1}^n\mathrm{DCG}_5^i, \]
where $\mathrm{DCG}_5^i = 1/\log_2{\left(p(i)+1\right)}$ and $p(i)$ is the position of the true destination in the list of five predicted destinations. For example, if for a particular user $i$ the destination is FR, and FR was at the top of the list of five predicted countries,  then 
\[
\mathrm{DCG}_5^i=\dfrac{1}{\log_2(1+1)}=1.0.
\] 
When FR is second, e.g. model prediction (US, FR, DE, NDF, IT) gives a 
\[
\mathrm{DCG}_5^i=\dfrac{1}{\log_2(2+1)}=1/1.58496=0.6309
\]

We trained our deep learning  network with 20 epochs and mini-batch size of 256.  For a hold-out sample we used 10\% of the data, namely 21346 observations. The fitting algorithm evaluates the $DCG$ function at every epoch to monitor the improvements of quality of predictions from epoch to epoch. It takes approximately 10 minutes to train, whereas the variational inference approach is computationally prohibitive at this scale. 

Our model uses a two-hidden layer architecture with ReLU activation functions
\begin{align*}
Y =  &\mathrm{softmax}(w^0Z^2 + b^0)\\
Z^2 = &\max(w^2Z^1 + b^2,0)\\
Z^1 = & \max(w^1X + b^1,0) \; .
\end{align*}
The weight matrices for simple data $W^1 \in \mathbb{R}^{64 \times p}$, $W^2 \in \mathbb{R}^{64 \times 64}$. In our notations, we assume that the activation function is applied point-vise at each layer.

The resulting model has out-of-sample $NDCG$ of  $-0.8351$. The classes are imbalanced in this problem. Table \ref{tab:dest_percent} shows percent of each class in out-of-sample data set.
\begin{table}
\centering
\begin{tabular}{c|c|c|c|c|c|c|c|c|c|c|c|c}
	Dest & AU & CA & DE & ES & FR & GB & IT & NDF & NL & PT & US & other \\ 
	\hline 
	\% obs & 0.3 & 0.6 & 0.5 & 1 & 2.2 & 1.2 & 1.2 & 59 & 0.31 & 0.11 & 29 & 4.8 \\ 
\end{tabular} 
\caption{Percent of each class in out-of-sample data set}
\label{tab:dest_percent}
\end{table}

\begin{figure}[H]
\centering
\includegraphics[width=0.5\textwidth]{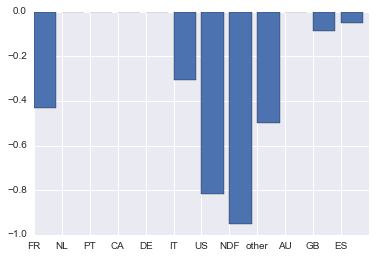}
\caption{Out-of-sample NDCG for each of the destinations}
\label{fig:ndcg}
\end{figure}

Figure \ref{fig:ndcg} shows out-of-sample NDCG for each of the destinations.

\begin{figure}
	\begin{tabular}{ccc}
	\includegraphics[width=0.33\textwidth]{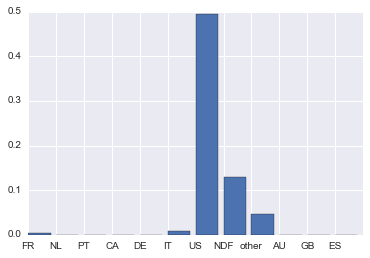} & \includegraphics[width=0.33\textwidth]{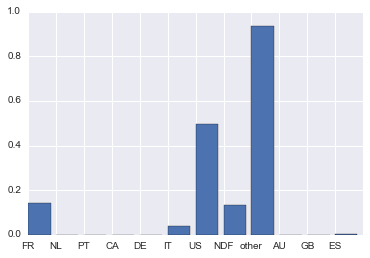} & \includegraphics[width=0.33\textwidth]{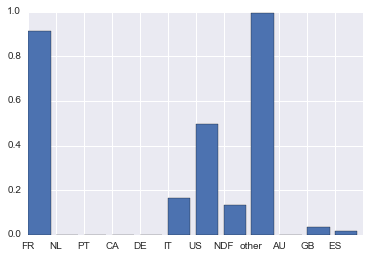}\\
	(a) first  & (b) second & (c) third
	\end{tabular}
	\caption{Prediction accuracy for deep learning  model. Panel (a) shows accuracy of prediction when only top predicted destination is used. Panel (b) shows correct percent of correct predictions when correct country is in the top two if the predicted list. Panel (c) shows correct percent of correct predictions when correct country is in the top three if the predicted list}
	\label{fig:pred_acc}
\end{figure}

Figure \ref{fig:pred_acc} shows accuracy of prediction for each of the destination countries. 
 The model accurately predicts bookings in the US and FR and other when top three predictions are considered. 
 
\begin{figure}[h!]
	\centering
	\includegraphics[width=0.8\textwidth]{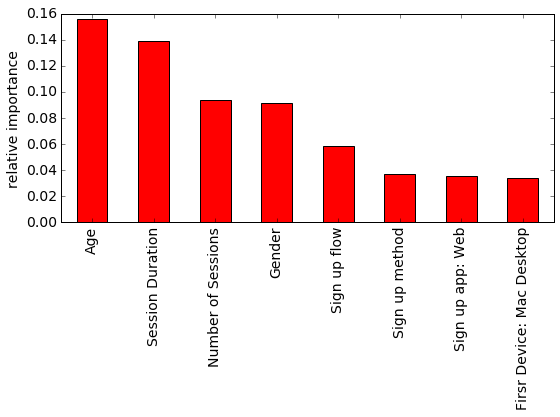}
%	\vspace{-15pt}
	\caption{The fifteen most important features as identified by the XGBoost model}
%	\vspace{-21pt}
	\label{fig:xgd_imp}
\end{figure}

Furthermore, we compared the performance of our deep learning model with the XGBoost algorithms \citep{chenG16} for fitting gradient boosted tree model. The performance of the model is comparable and yields NGD of $-0.8476$. One of the advantages of the tree-based model is its  ability to calculate the importance of each of the features \citep{hastie_elements_2016}. Figure \ref{fig:xgd_imp} shows the variable performance calculated from our XGBoost model.

The importance scores calculated by the  XGBoost model confirm our exploratory data analysis findings. In particular, we see the fact that a user specified gender is a strong predictor. Number of sessions on Airbnb site recorded for a given user before booking is a strong predictor as well. Intuitively, users who visited the site multiple times are more likely to book. Further,  web-users who signed up via devices with large screens are also likely to book as well.

\section{Discussion}
Our view of deep learning is a high dimensional nonlinear data reduction scheme, generated probabilistically  as a stacked generalized linear model (GLM). This sheds light on how to  train a deep architecture using SGD. This is a first order gradient method for finding a posterior  mode in a very high dimensional space. By taking a predictive approach, where regularization learns the architecture, deep learning has been very successful in many fields.  

There are many areas of future research for Bayesian deep learning which include
\begin{enumerate}[label=(\roman*)]
	\item By viewing deep learning probabilistically as stacked GLMs allows many statistical models such as exponential family models and heteroscedastic errors.
	\item  Bayesian hierarchical models have similar advantages to deep learners. Hierarchical models include extra stochastic layers and provide extra interpretability and flexibility. 
	%	\item Another avenue is combining proximal algorithms and MCMC. 
	\item By viewing deep learning as a Gaussian Process allows for exact Bayesian inference \cite{neal1996priors, williams1997computing,lee2017deep}. The Gaussian Process connection opens opportunities to develop more flexible and interpretable models for engineering \cite{gramacy2011particle} and natural science applications \cite{banerjee2008gaussian}.
	\item With gradient information easily available via the chain rule (a.k.a. back propagation), a new avenue of stochastic methods to fit networks exists, such as MCMC, HMC, proximal methods, and ADMM, which could dramatically speed up the time to train deep learners.
	%	\item  Hyper-parameter tuning 
	\item Comparison with traditional Bayesian non-parametric approaches, such as treed Gaussian Models \citep{gramacy2005bayesian}, and BART \citep{chipman2010bart} or using hyperplanes in Bayesian non-parametric methods ought to yield good predictors \citep{bass2017}. 
	\item Improved Bayesian algorithms for hyper-parameter training and optimization \citep{snoek2012practical}. Langevin diffusion MCMC, proximal MCMC and Hamiltonian Monte Carlo (HMC) can exploit the derivatives as well as Hessian information \citep{polson_proximal_2015,polson2015statistical,dean_large_2012}.
	\item Rather than searching a grid of values with a goal of minimising out-of-sample means squared error, one could place further regularisation penalties (priors) on these parameters and integrate them out.
\end{enumerate}
MCMC methods also have lots to offer to DL and can be included seamlessly in \verb|TensorFlow| \citep{tensorflow2015-whitepaper}. Given the availability  of high performance computing, it is now possible to implement high dimensional posterior inference on large data sets is now a possibility, see \cite{dean_large_2012}. The same advantages are now available for Bayesian inference.  Further, we believe deep learning models have a bright future in many fields of applications, such as finance, where DL is a form of nonlinear factor models \citep{heaton2016deep,heaton2016deepa}, with each layer capturing different time scale effects and spatio-temporal data is viewed as an image in space-time \citep{dixon2017,polson_deep_2017}. In summary, the Bayes perspective adds helpful interpretability, however, the full power of a Bayes approach has still not been explored. From a practical perspective, current regularization approaches have provided great gains in predictive model power for recovering nonlinear complex data relationships. 

%\vnote{Add Gel reference and Edward reference (maybe)}
\bibliography{ref}
\ifarxiv
	\bibliographystyle{alpha}
\else
	\bibliographystyle{ba}
\fi
\vspace{0.3in}

\end{document}